\documentclass[11pt]{article}

\usepackage[final]{acl}

\usepackage{times}
\usepackage{latexsym}

\usepackage[T1]{fontenc}

\usepackage[utf8]{inputenc}
\usepackage{kotex}

\usepackage{microtype}

\usepackage{inconsolata}

\usepackage{graphicx}
\usepackage{stfloats}

\usepackage{comment}
\usepackage{algorithm}
\usepackage{algpseudocode}

\usepackage{array}
\usepackage{booktabs}
\usepackage{tabularx}
\newcolumntype{L}{>{\raggedright\arraybackslash}X}
\usepackage{multirow}
\usepackage{amsmath}
\usepackage{amssymb}
\usepackage{multirow}
\usepackage{booktabs}
\usepackage{tabularx}
\newcolumntype{L}{>{\raggedright\arraybackslash}X}
\usepackage{fvextra}
\usepackage{mdframed}
\usepackage{listings}
\mdfdefinestyle{promptbox}{
  linewidth=0.5pt,
  innerleftmargin=8pt,
  innerrightmargin=8pt,
  innertopmargin=5pt,
  innerbottommargin=5pt,
  backgroundcolor=gray!5,
}

\newcommand{\jh}[1]{\textcolor{black}{#1}}

\title{LandingAgent: A Reference-Annotated Dataset and Agentic Generation Framework for Landing Pages}

\author{
  Injun Baek\textsuperscript{1,2,*} \quad
  HyeongSeok Lee\textsuperscript{1,*} \quad
  Yearim Kim\textsuperscript{1} \quad
  Junhoo Lee\textsuperscript{1} \quad
  Nojun Kwak\textsuperscript{1,\textdagger}
  \\[0.35em]
  {\normalfont
    \textsuperscript{1}Seoul National University \quad
    \textsuperscript{2}Samsung Electronics
  }
  \\
  {\normalfont\texttt{\{jjune1416, tjdgns2048, yerim1656, mrjunoo, nojunk\}@snu.ac.kr}}
  \\[-0.1em]
  {\normalfont\small
    \textsuperscript{*}Equal contribution. \quad
    \textsuperscript{\textdagger}Corresponding author.
  }
}

\makeatletter
\AtBeginDocument{%
\def\@maketitle{\vbox to \titlebox{\hsize\textwidth
 \linewidth\hsize \vskip 0.125in minus 0.125in \centering
 {\Large\bfseries \@title \par} \vskip 0.2in plus 1fil minus 0.1in
 {\def\and{\unskip\enspace{\rmfamily and}\enspace}%
  \def\And{\end{tabular}\hss \egroup \hskip 1in plus 2fil
           \hbox to 0pt\bgroup\hss \begin{tabular}[t]{c}\bfseries}%
  \def\AND{\end{tabular}\hss\egroup \hfil\hfil\egroup
          \vskip 0.25in plus 1fil minus 0.125in
           \hbox to \linewidth\bgroup\large \hfil\hfil
             \hbox to 0pt\bgroup\hss \begin{tabular}[t]{c}\bfseries}
  \hbox to \linewidth\bgroup\large \hfil\hfil
    \hbox to 0pt\bgroup\hss
  \outauthor
   \hss\egroup
    \hfil\hfil\egroup}
  \vskip 0.18in
  \includegraphics[width=\textwidth]{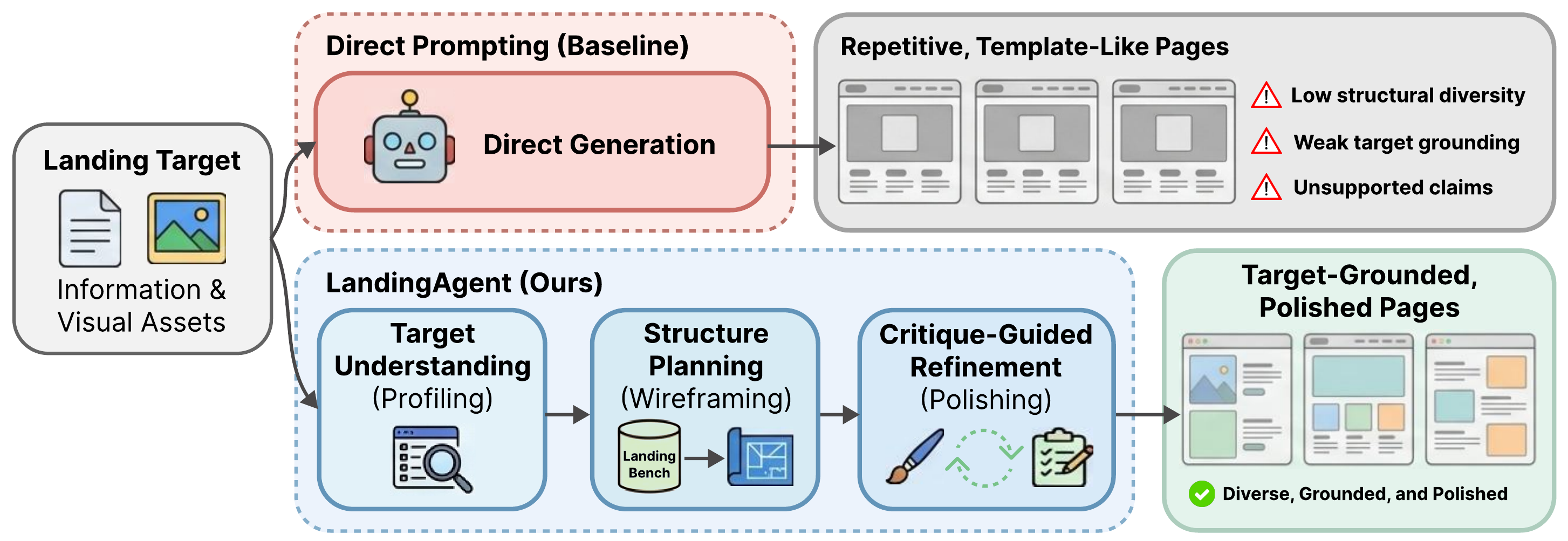}\par
  \vspace{-0.06in}
  {\captionsetup{hypcap=false}
  \captionof{figure}{
Overview of target-grounded, reference-guided landing-page generation.
Direct Prompting can produce repetitive, template-like pages with weak target grounding and unsupported claims.
\textbf{LandingAgent} addresses these failures by profiling the target, adapting \textbf{LandingBench} reference profiles during wireframing, and refining the draft through critique-guided polishing, yielding target-grounded, structurally diverse, and visually polished pages.
}
  \label{fig:teaser}}
  \vfil
}}
}
\makeatother

\begin{document}
\maketitle
\begin{abstract}

Landing pages are goal-oriented web interfaces that must communicate a target-specific value proposition while organizing information flow, visual hierarchy, and calls to action (CTA). Although large language models can generate plausible webpage code from natural-language prompts, direct generation often yields generic templates and unsupported persuasive claims. We study target-grounded, reference-guided landing-page generation, where a system must create an executable page for a new target by adapting reusable patterns from real pages without copying them. We introduce LandingBench, a reference-profile dataset that abstracts real landing pages into section sequences, layout patterns, tone descriptors, visual emphasis, and CTA structure. Building on LandingBench, we propose LandingAgent, a three-phase agentic framework that profiles the target, constructs a reference-guided wireframe, and refines the page through critique-guided polishing. We evaluate LandingAgent against direct prompting on faithfulness, conciseness, readability, aesthetics, and structural diversity. Experiments show improved target grounding, presentation quality, and layout diversity. Code is available at \url{https://github.com/IAURAI/LandingAgent}.


\end{abstract}

\section{Introduction}
\label{sec:intro}

Landing pages are conversion-oriented web interfaces that translate a product, service, project, or organization into a concise user experience. They shape how users first understand a target's value proposition and whether they take actions such as sign-up, purchase, subscription, or demo request. Generating an effective landing page therefore requires more than executable webpage code; it requires coordinated decisions about message selection, information ordering, visual hierarchy, tone, trust cues, and call-to-action (CTA) placement.

Recent large language models (LLMs) have made it increasingly feasible to generate webpages from natural-language prompts. Given a short description of a product or service, an LLM can produce plausible HTML, CSS, and JavaScript, substantially reducing the cost of web prototyping. However, landing-page generation is not merely a code synthesis problem. Direct LLM generation may repeatedly produce similar template structures~\cite{rel2024does} and provides limited control over target-specific information architecture. It may also introduce unsupported persuasive claims, such as non-existent features, customer evidence, performance numbers, or business outcomes~\cite{rel2025hallucination}. These failures are particularly problematic in landing pages, where visually salient copy can amplify information that is not grounded in the input specification.

Existing landing pages offer valuable reference signals, since they show how real products organize value propositions, section flows, visual hierarchies, and CTAs. However, raw pages are a poor conditioning format. Their reusable design and communication strategies are entangled with source-specific copy, branding, visual assets, and implementation details. Providing full HTML or complete page content to an LLM can therefore introduce noise and increase the risk of surface-level imitation rather than target-specific adaptation. We study \textbf{target-grounded, reference-guided landing-page generation}: generating a new executable landing page for a given target by adapting abstract reference knowledge while avoiding source-specific copying and unsupported persuasive content. Figure~\ref{fig:teaser} illustrates this setting.

To support this, we introduce \textbf{LandingBench}, a reference-profile dataset constructed from real-world landing pages. Rather than treating collected pages as templates, LandingBench abstracts each page into reusable annotations, including section sequences, layout patterns, visual emphasis strategies, tone descriptors, CTA structures, and information-density attributes. These profiles allow a generation system to retrieve references that are structurally and rhetorically relevant to a target specification while reducing dependence on the source page's full HTML, copy, or brand-specific assets.

Building on LandingBench, we propose \textbf{LandingAgent}, a three-phase agentic framework for landing-page generation. The \emph{Profiling} phase converts heterogeneous user inputs into a Structured Page Brief that serves as the grounding source for subsequent steps. The \emph{Wireframing} phase retrieves a relevant reference profile and constructs a target-specific page structure, separating reference adaptation from direct template reuse. The \emph{Polishing} phase refines the approved structure into a final executable landing page by improving copy, visual presentation, and functional consistency.
We evaluate LandingAgent against a Direct Prompting baseline and show that reference-guided agentic generation improves target faithfulness, presentation quality, and structural diversity.

Our contributions are threefold:
\begin{itemize}
    \item We introduce \textbf{LandingBench}, a reference-profile dataset abstracting real-world landing pages into reusable structural, rhetorical, and visual annotations rather than raw templates.
    \item We propose \textbf{LandingAgent}, a three-phase framework decomposing landing-page generation into Profiling, reference-guided Wireframing, and critique-guided Polishing.
    \item We evaluate LandingAgent against direct LLM generation across target faithfulness, presentation quality, and structural diversity. 
\end{itemize}

\providecommand{\jh}[1]{\textcolor{purple}{#1}}

\section{Related Work}

\subsection{LLM-based Code and Web Generation}
Recent LLMs have substantially improved the ability to synthesize executable code from natural-language requirements, and this progress has extended beyond general programming tasks to web-interface generation~\cite{rel2024uicoder2024, rel2026webgen, rel2025design2code}.
Prior work on LLM-based code generation has primarily focused on generating syntactically valid and functionally correct code from user specifications.
Code-specialized models such as Codex~\cite{rel2021codex} and Code Llama~\cite{rel2023codelama} have shown that LLMs can perform practical code-writing tasks across diverse programming languages and scenarios.

Landing-page generation, however, is not merely specification-to-code translation.
It requires organizing a target's value proposition, information flow, visual hierarchy, and CTA strategy into a coherent and persuasive page.
Because direct prompting can lead to repetitive templates~\cite{rel2024does} or unsupported claims~\cite{rel2025hallucination}, we frame the task as a goal-oriented web generation problem requiring target faithfulness, presentation quality, and structural diversity.


\subsection{Datasets for UI and Webpage Generation}
Datasets and benchmarks for UI and webpage generation have also been actively studied.
Early work such as pix2code~\cite{rel2018pix2code} introduced the task of generating platform-specific UI code from GUI screenshots. More recent datasets have scaled this direction to web interfaces. WebSight~\cite{rel2024webSight} presents a large-scale dataset of webpage screenshots paired with HTML code, while Web2Code~\cite{rel2024web2code} evaluates HTML generation and webpage understanding from webpage images and instructions. Design2Code~\cite{rel2025design2code} further studies how accurately a model can reproduce frontend code from a given visual input based on real webpages.
These datasets primarily evaluate whether a model can faithfully reconstruct a given visual interface as code. However, landing-page generation poses a different challenge: the model must plan how to organize a new target into a persuasive page structure, including section order, information flow, layout patterns, tone, and CTA composition. We therefore construct LandingBench as a reference dataset that captures reusable communication and design patterns from real landing pages through abstract annotations rather than screenshot--code pairs.

\subsection{Agent/Retrieval-based Web Generation}
Retrieval-augmented generation (RAG)~\cite{rel2020RAG} and agent-based methods~\cite{rel2024agentsurvey} have been explored to improve reliability and controllability in complex generation tasks.
ReAct~\cite{rel2022react} interleaves reasoning and action to support interaction with external environments or knowledge sources, whereas Reflexion~\cite{rel2023reflexion} uses linguistic feedback to improve subsequent trials.
Recent agentic systems also highlight the value of standardized agent-data representations, specialized roles, and fine-grained evaluation protocols for complex generative workflows~\cite{rel2025agent, rel2025coda, rel2025edival}.

In web generation, WebGen-Bench evaluates multi-file websites using functional tests executed by a navigation agent, whereas WebGen-Agent iteratively refines websites through screenshot feedback~\cite{rel2026webgen,rel2025webgenagent}.
MM-WebAgent uses hierarchical planning to generate multimodal webpages with AI-generated visual elements~\cite{rel2026mmwebagent}.
These systems are closely related but use different task formulations and evaluation settings, complicating direct comparison.
Our task instead focuses on generating a single target-grounded landing page from a product, service, or project specification.
In this work, we retrieve abstract landing-page annotations rather than raw webpage content and organize generation into three phases: Profiling, Wireframing, and Polishing.
This design separates reference-guided structural planning from local refinement of copy, visuals, and functionality.

\begin{figure}[t]
    \centering
    \includegraphics[width=.8\columnwidth]{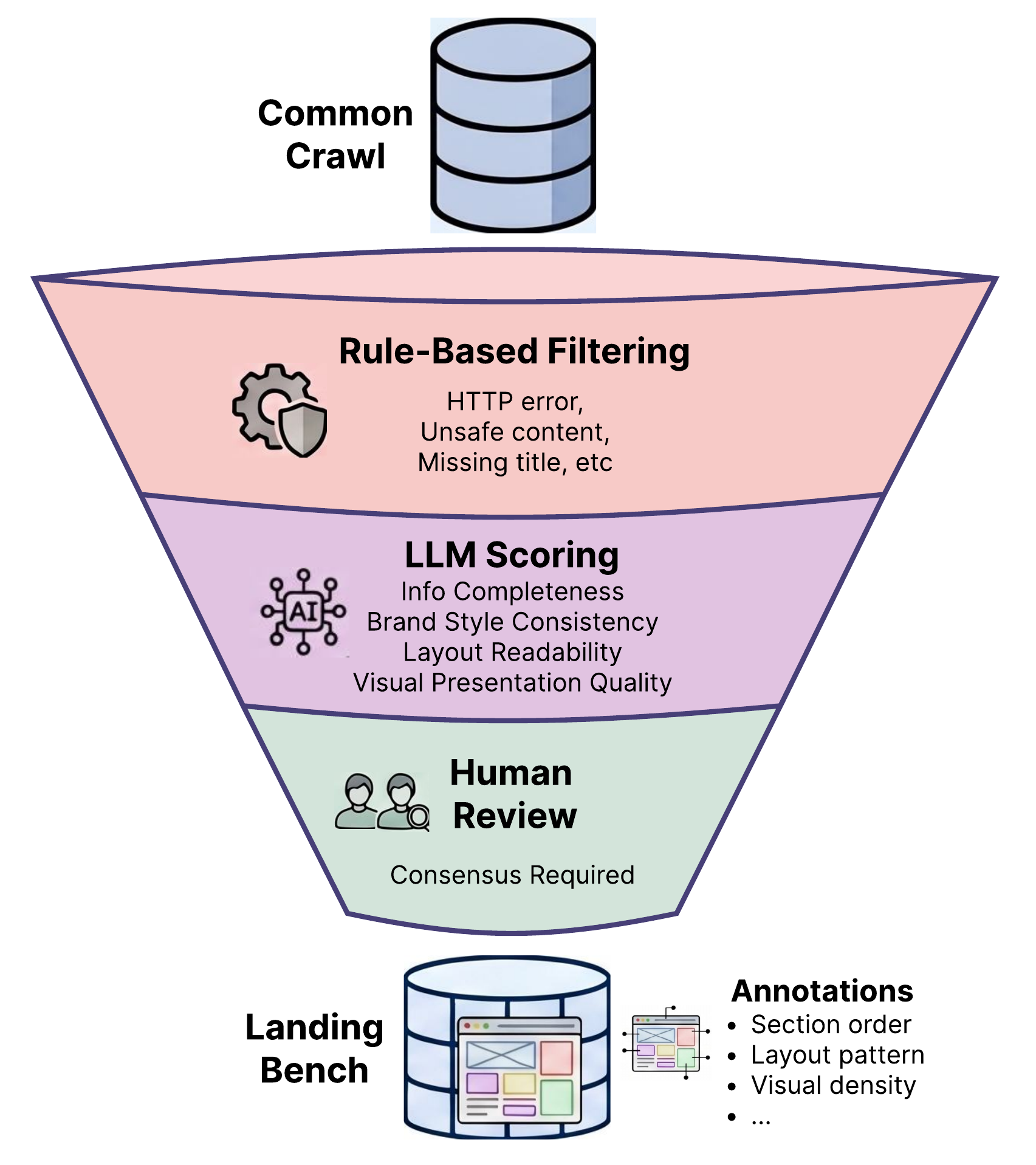}
    \caption{Overview of the LandingBench collection and filtering pipeline. Candidate pages sampled from Common Crawl are processed through rule-based filtering, LLM-based scoring and human verification.} 
    \label{fig:dataset-pipeline}
\end{figure}

\section{LandingBench: A Landing-Page Reference Annotation Dataset}
\label{sec:landingbench}

LandingBench is designed to make real-world landing pages usable as abstract reference knowledge for target-grounded generation. Rather than treating collected pages as templates to be copied, LandingBench represents each page through a compact reference profile that captures reusable structural, rhetorical, and visual patterns. These profiles describe how a page organizes information, presents its value proposition, establishes visual emphasis, and guides users toward conversion actions.

This design distinguishes LandingBench from conventional UI and webpage generation datasets, which primarily support visual or code-level reconstruction. In our setting, the objective is not to reproduce a given screenshot or source implementation. Instead, a generation system must retrieve relevant reference patterns and adapt them to a new landing target without copying the source page's copywriting, branding, layout, or implementation artifacts. LandingBench therefore focuses on the communicative structure of landing pages: how real pages sequence sections, shape information flow, express tone, and position calls to action.
More detailed information is provided in Appendix~\ref{app:landingbench-details}.

\begin{table*}[t]
\centering
\small
\begin{tabularx}{\textwidth}{lX}
\toprule
\textbf{Level} & \textbf{Reference-profile fields} \\
\midrule
Page-level axes
& category, target, conversion goal, tone, information density, scroll depth, CTA, asset dependency \\
Section structure
& section order, section role, layout pattern, visual emphasis \\
Visual descriptors
& palette descriptor, typography descriptor, imagery \& iconography descriptor \\
Retrieval fields
& page-structure axes tuple, reference screenshot, abstract metadata \\
\bottomrule
\end{tabularx}
\caption{Core reference-profile schema of LandingBench.}
\label{tab:reference-profile-schema}
\end{table*}

\subsection{Collection and Filtering}

We collect landing-page candidates from Common Crawl~\cite{commoncrawl}, a large-scale public web archive.
From the initial candidate pool, we randomly sample 1,000 pages and process them through a three-stage quality-control pipeline, as in Figure~\ref{fig:dataset-pipeline}. 
The goal of this pipeline is not merely to select visually polished pages, but to identify pages whose structural patterns and communication strategies can be reliably annotated and later used as references for retrieval-based generation.

\paragraph{Hard Filtering.}
The first stage removes clearly unsuitable pages for LandingBench. We combine rule-based checks with LLM-assisted eligibility judgments using \texttt{Qwen3.6\_35B\_A3B}~\cite{bai2025qwen3}. Rule-based checks remove inaccessible/unstable pages, including pages with HTTP errors, screenshot-capture failures, login requirements, bot verification, region restrictions, or request blocking. We also exclude pages whose URL patterns or visible text indicate non-landing-page formats, such as blogs, news articles, job postings, documentation pages, help pages, support pages, wikis, and product-detail pages. Pages containing research-unsuitable material are also removed. The LLM-assisted eligibility judgment further checks whether each candidate is accessible, landing-page-like, sufficiently informative, visually interpretable, legitimate, and annotatable. We then remove near-duplicate pages using perceptual hashing on screenshots and TF-IDF similarity on visible text, retaining the page with the higher soft-filter total score.

\paragraph{LLM-based Scoring.}
The second stage estimates whether each remaining page is suitable as a reusable landing-page reference. We use \texttt{Qwen3.6\_35B\_A3B} 
with each page's full-page screenshot and complete visible text as input. The model assigns 1--5 scores on four dimensions: \textit{Information Completeness}, which measures whether the page sufficiently presents product or service information and a value proposition; \textit{Brand-Style Consistency}, which measures whether tone and visual style remain coherent; \textit{Layout Readability}, which measures whether section boundaries and information flow are clear; and \textit{Visual Presentation Quality}, which measures overall visual completeness and polish. Only pages scoring 4 or higher on all four dimensions are passed to human verification. These scores are used only for candidate selection, not as gold-standard quality labels for evaluation.

\paragraph{Human Review.}
The final stage consists of independent human verification by two reviewers.
Each reviewer inspects the original webpage, full-page screenshot, and complete visible text twice. Reviewers assess whether the candidate is a genuine landing page and whether its section structure can be reliably identified. 
A page is included in LandingBench only when both reviewers assign an \textsc{Accept} decision, ensuring that the final dataset is both reference-suitable, in the sense that the page contains reusable design, 
and annotation-suitable, in the sense that its structure can be consistently represented as a reference profile.
Across 636 candidates, agreement between the two reviewers was $P_o=0.821$, with Cohen's $\kappa=0.48$.


\subsection{Reference Profile Annotation}

Each verified page is converted into a compact reference profile, a structured representation that exposes reusable landing-page patterns without providing the source page's full HTML or complete verbatim copy as generation material. Each profile contains page-level axes, section-level structure, visual descriptors, and retrieval fields (Table~\ref{tab:reference-profile-schema}).

At the page level, we annotate attributes that characterize the overall communicative and conversion context of the page, including product or service category, target audience, conversion goal, tone, information density, scroll depth, CTA structure, and asset dependency. At the section level, we annotate the section order, section role, layout pattern, and visual emphasis. These fields allow a generation model to reason about how the page organizes its rhetorical flow, such as moving from a hero section to feature explanation, social proof, pricing, FAQ, or final CTA. At the visual level, we store abstract descriptors for palette, typography, and imagery \& iconography. The reference screenshot is retained as visual context, but the profile is designed to discourage direct reuse of source-specific design elements.

This abstraction is central to LandingBench.
Raw HTML is often long, noisy, and implementation-specific, and complete page copy may contain source-specific claims, slogans, or brand language. 
By representing pages through reference profiles, LandingBench provides generation systems with structured guidance about landing-page organization while reducing the risk of surface-level copying.

\subsection{Dataset Statistics}

After filtering and verification, LandingBench contains 438 pages covering 13 section-role labels, with average 8.0 annotated sections per page.

\section{LandingAgent}
\label{sec:method}
\begin{figure*}[!t]
    \centering
    \includegraphics[width=\textwidth]{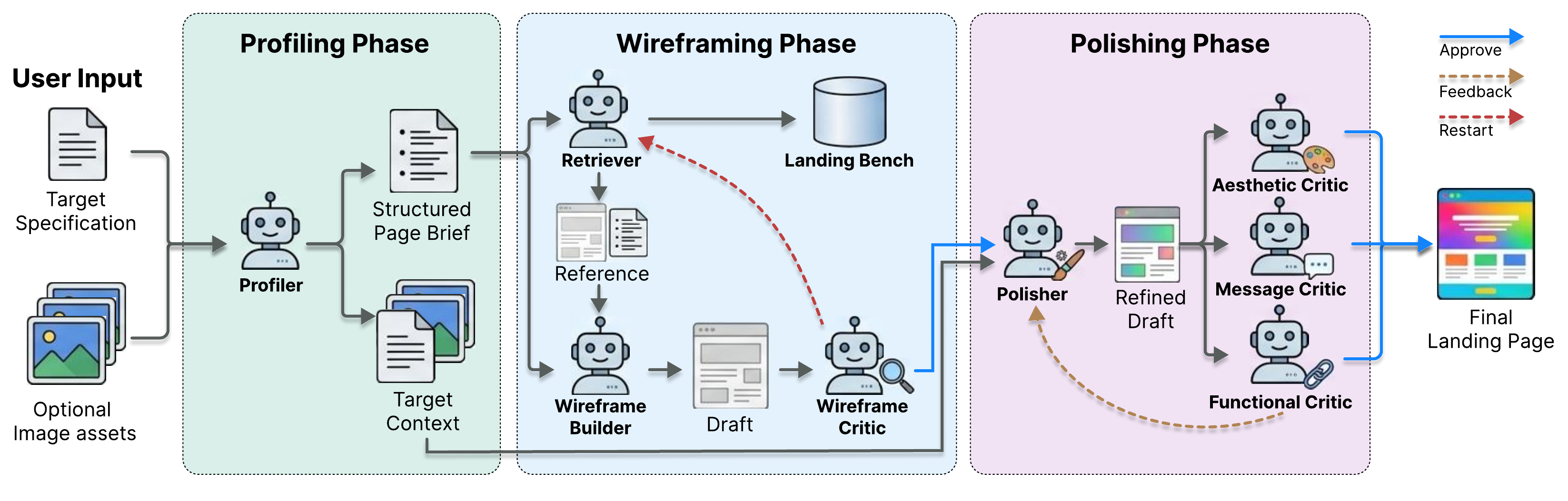}
    \caption{Overview of the LandingAgent framework. Given a user-provided target specification and optional image assets, LandingAgent first performs a Profiling Phase that transforms the inputs into a Structured Page Brief. In the Wireframing Phase, a Retriever selects a structurally relevant reference from LandingBench, and a WireframeBuilder constructs a skeletal page draft. A WireframeCritic either approves the draft or triggers a Restart Loop that reselects a reference and regenerates the wireframe. In the Polishing Phase, a Polisher iteratively refines the draft with feedback from three parallel critics: AestheticCritic, MessageCritic, and FunctionalCritic.} 
    \label{fig:landing-agent-overview}
\end{figure*}
\begin{table*}[t]
\centering
\small
\renewcommand{\arraystretch}{1.18}
\begin{tabularx}{\textwidth}{@{}ll>{\raggedright\arraybackslash}X@{}}
\toprule
\textbf{Phase} & \textbf{Agent} & \textbf{Role} \\
\midrule
Profiling
& \textsc{Profiler}
& Transforms the target specification and image assets into a Structured Page Brief. \\

\midrule
Wireframing
& \textsc{Retriever}
& Selects a reference profile from LandingBench that matches the Structured Page Brief along structural and visual axes. When a restart occurs, it uses the diagnostic hint to avoid repeating the same structural failure. \\

& \textsc{WireframeBuilder}
& Constructs a skeletal page draft guided by the selected reference profile at the level of section roles, information flow, and layout organization, without using raw HTML or verbatim source-page copy. \\

& \textsc{WireframeCritic}
& Evaluates whether the draft is globally aligned with the Structured Page Brief and the retrieved reference profile. It either approves the draft for polishing or triggers a restart with a diagnostic hint. \\

\midrule
Polishing
& \textsc{Polisher}
& Refines the approved wireframe into an executable landing page by adding target-grounded copy, visual styling, asset placement, and implementation details while preserving the global structure. \\

& \textsc{AestheticCritic}
& Assesses visual presentation quality, including typography hierarchy, color harmony, whitespace, alignment, visual readability, modernness, brand expression and reference fidelity. \\

& \textsc{MessageCritic}
& Checks whether the generated page faithfully reflects the Structured Page Brief along five aspects: tone fidelity, value-proposition clarity, audience signaling, must-have-section coverage, and claim grounding. \\

& \textsc{FunctionalCritic}
& Audits implementation-level consistency, including anchor integrity, navigation coverage, and CTA targeting. \\

\bottomrule
\end{tabularx}
\caption{Functional role of each agent, grouped by phase.}
\label{tab:agent_roles}
\end{table*}
\textbf{LandingAgent} is a three-phase framework for generating executable landing pages from a target specification and optional image assets.
As shown in Figure~\ref{fig:landing-agent-overview}, the framework first transforms the input into a Structured Page Brief, then constructs a reference-guided wireframe using LandingBench, and finally refines the draft through critique-guided polishing. 
The design separates two types of failures: global structural misalignment, which is handled by restarting the wireframing process, and local textual, visual, or functional defects, which are handled through iterative polishing.
Table~\ref{tab:agent_roles} summarizes agent roles in LandingAgent.
See Appendix~\ref{app:landingagent-details} for details, including prompt templates.

\subsection{Profiling}

The Profiling phase transforms heterogeneous user inputs into a Structured Page Brief, which serves as the grounding source for retrieval, generation, and critique. The brief contains the target's value proposition, intended audience, product or service category, tone descriptors, conversion goal, required information, and visual identity hints. It also includes retrieval axes aligned with LandingBench, such as information density, expected scroll depth, CTA structure, and asset dependency. This transformation allows subsequent agents to operate on a consistent target representation rather than directly conditioning on raw user input.

\subsection{Reference-Guided Wireframing}

The Wireframing phase determines the global structure of the landing page. The Retriever selects a reference profile from LandingBench using the retrieval axes in the Structured Page Brief. The WireframeBuilder 
generates a page draft conditioned on the brief and the selected reference profile.

The reference is not provided as raw HTML or verbatim source-page copy. Instead, LandingAgent conditions on an abstract reference profile containing section structure, layout descriptors, visual descriptors, and retrieval metadata. This representation exposes reusable landing-page patterns while reducing implementation-level noise and the risk of direct reference copying.

After each draft is generated, the WireframeCritic checks whether the page structure aligns with the Structured Page Brief and whether the reference has been used as abstract guidance rather than as a template. If approved, the draft proceeds to polishing. Otherwise, the critic returns a diagnostic hint, which is appended to the next retrieval attempt. The retrieve--build--critique process repeats up to $R_{\max}=3$ times; once the restart limit is reached, the most recent draft is passed to the Polishing phase.

\subsection{Critique-Guided Polishing}

The Polishing phase refines the approved wireframe into a final landing-page implementation. Unlike the Wireframing phase, it preserves the section order and major layout structure, focusing on copy clarity, CTA wording, typography, spacing, visual coherence, and implementation consistency.

Polishing iterates between the Polisher and three parallel critics.
The AestheticCritic evaluates visual presentation, the MessageCritic evaluates target faithfulness and copy quality, and the FunctionalCritic evaluates one-page implementation consistency.
Each critic returns a score, an approval decision, and feedback, including a structured patch when applicable. A page is accepted only when all critics meet the approval threshold, set to 4 out of 5 by default. Otherwise, the feedback is passed back to the Polisher, and refinement continues until all critics approve or the maximum number of polishing rounds $T_{\max}=3$ is reached.

\section{Evaluation}
\label{sec:evaluation}

\subsection{Setup}

We compare LandingAgent with zero-shot \textit{Direct Prompting}, \textit{One-shot Prompting} using one in-context example, and \textit{Direct Prompting + Polisher}, which applies LandingAgent's Polishing phase to a direct output.
All methods use the same 30 target specifications, output requirements, and evaluation metrics.
One-shot Prompting uses no critic, whereas Direct Prompting + Polisher shares LandingAgent's generator, critic, and three-round polishing limit.
Token counts are not matched because the methods have different numbers of stages.

We evaluate structural diversity, content quality (\textit{Faithfulness}, \textit{Conciseness}), and presentation quality (\textit{Readability}, \textit{Aesthetics}); the latter four criteria are adapted from~\citet{metric2018aesthetics}.
We compute Diversity using DINOv3~\cite{eval2025dinov3} representations and score the other metrics with LLM-as-a-Judge (LaaJ) on a 1--5 scale, supplemented by a human study.
Full metric definitions and judge prompts are provided in Appendix~\ref{app:evaluation_details}.

The 30 targets span 12 industries and vary in audiences, CTA goals, and page styles, with category-level references in LandingBench (see Appendix~\ref{app:evaluation_target_coverage}).
We generate three pages per target and configuration, for 540 pages in total.

\begin{table*}[t]
    \centering
    \small
    \setlength{\tabcolsep}{2.1pt}
    \begin{tabular}{@{}lll|>{\centering\arraybackslash}m{0.55in}|>{\centering\arraybackslash}m{0.70in}>{\centering\arraybackslash}m{0.70in}|>{\centering\arraybackslash}m{0.70in}>{\centering\arraybackslash}m{0.70in}@{}}
        \toprule
        \multirow{2}{*}{Method} & \multirow{2}{*}{Generator} & \multirow{2}{*}{Critic} & \multirow{2}{*}{Diversity} & \multicolumn{2}{c|}{Content} & \multicolumn{2}{c}{Presentation} \\
        & & & & Faithfulness & Conciseness & Readability & Aesthetics \\
        \midrule
        \multirow{2}{*}{\shortstack[l]{Direct\\Prompting}} & gemini-2.5-pro & --- & 0.26 & 3.83 & 4.40 & 3.90 & 3.67 \\
         & gpt-5 & --- & 0.28 & 4.60 & 4.13 & 3.77 & 3.63 \\
        \midrule
        \shortstack[l]{One-shot\\Prompting} & gemini-2.5-pro & --- & 0.29 & 4.05 & 4.67 & 3.95 & 3.67 \\
        \midrule
        \shortstack[l]{Direct Prompting\\+ Polisher} & gemini-2.5-pro & claude-sonnet-4-6 & 0.25 & 4.33 & 4.71 & 3.95 & 3.62 \\
        \midrule
        \multirow{2}{*}{\shortstack[l]{Landing\\Agent}} & gemini-2.5-pro & claude-sonnet-4-6 & 0.36 & 4.30 & \textbf{4.83} & 4.03 & 4.00 \\
         & gpt-5 & claude-sonnet-4-6 & \textbf{0.37} & \textbf{4.70} & 4.50 & \textbf{4.07} & \textbf{4.10} \\
        \bottomrule
    \end{tabular}
    \caption{Main evaluation results with structural diversity, content quality (\textit{Faithfulness}, \textit{Conciseness}), and presentation quality (\textit{Readability}, \textit{Aesthetics}). 
    Scores are rounded to two decimal places.}
    \label{tab:main-evaluation-results}
\end{table*}

\begin{figure*}[t]
    \centering
    \includegraphics[width=\textwidth]{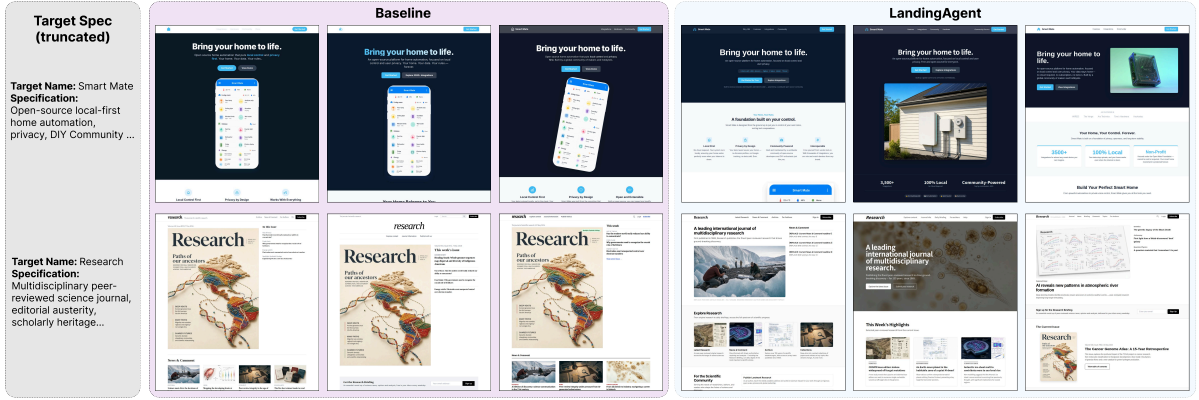}
    \caption{Qualitative comparison of structural diversity. For the same target specifications, Direct Prompting repeats hero layouts, visual assets, and section order, whereas LandingAgent produces diverse yet target-faithful layouts, visual grounding, and section flows.}
    \label{fig:diversity-qualitative}
\end{figure*}

\subsection{Main Results}

Table~\ref{tab:main-evaluation-results} shows One-shot Prompting improves overall over Direct Prompting, with Aesthetics unchanged.
Direct Prompting+Polisher improves Faithfulness and Conciseness but reduces Diversity and Aesthetics.

LandingAgent outperforms One-shot Prompting on all five metrics and Direct Prompting + Polisher on all but Faithfulness, where scores are comparable.
Together with the \textit{w/o RAG} and \textit{w/o Profiling} ablations in Table~\ref{tab:ablation-results}, these comparisons separate the effects of prompting, polishing, reference retrieval, and Profiling with reference-guided Wireframing.

\begin{figure}[t]
    \centering
    \includegraphics[width=\columnwidth]{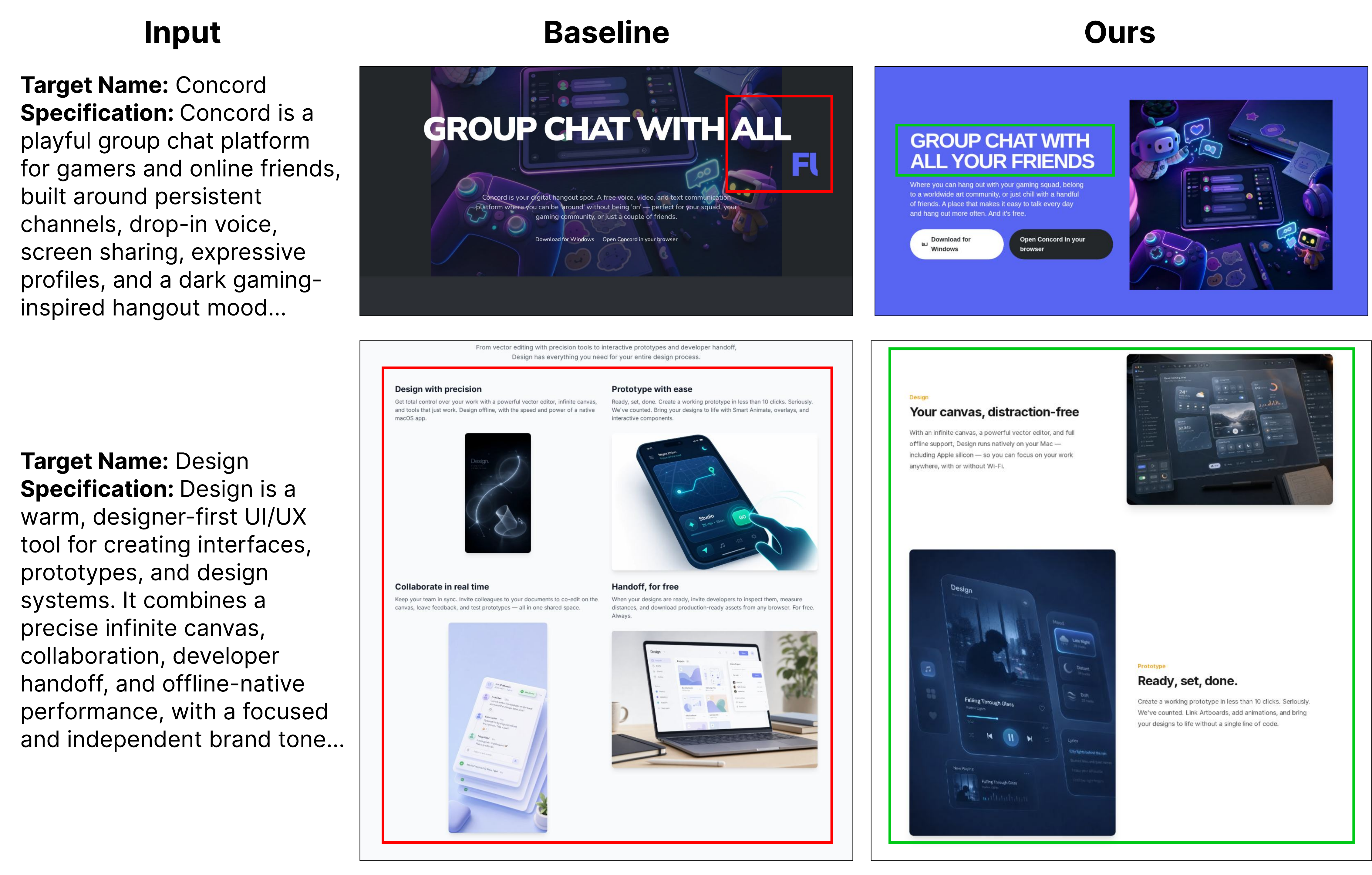}
    \caption{Qualitative comparison of Direct Prompting and LandingAgent on two target specifications: Faithfulness (top) and Aesthetics (bottom). Red boxes mark failure regions; green boxes mark successful regions.}
    \label{fig:main-qualitative}
\end{figure}

Figure~\ref{fig:diversity-qualitative} shows that Direct Prompting repeats layouts, whereas LandingAgent varies section flows while preserving target relevance.
In the Faithfulness example at the top of Figure~\ref{fig:main-qualitative}, the baseline's hero headline overflows its container and is clipped beyond the visible region, obscuring target-grounded messaging, whereas LandingAgent renders the full headline.
On Aesthetics (bottom), the baseline arranges visual assets with inconsistent sizing and misaligned spacing across a dense grid, producing a visually cluttered composition, whereas LandingAgent yields a cleaner layout with well-aligned imagery and balanced whitespace.
See Appendix~\ref{app:more_qualitative} for further qualitative comparisons.

\subsection{User Study}
\begin{figure}[t]
    \centering
    \includegraphics[width=\columnwidth]{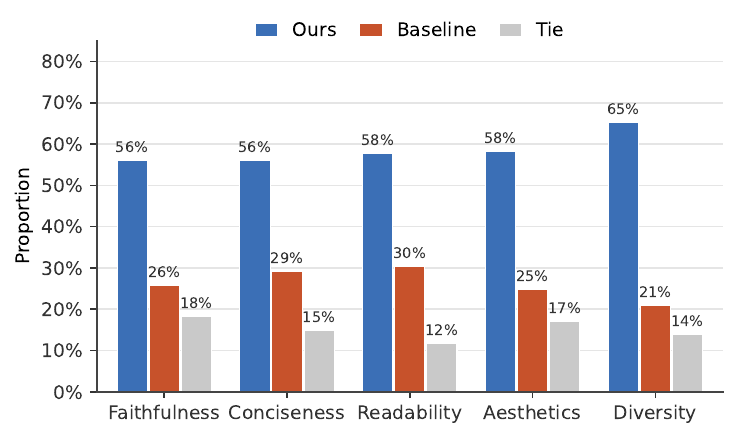}
    \caption{Pairwise user study results: Ours (LandingAgent) vs.\ Baseline preferences across the five criteria.}
    \label{fig:userstudy}
\end{figure}

We conduct a blind pairwise user study as a human evaluation.
Participants are shown two landing pages (Ours, Baseline) generated from the same source side-by-side, and for each of five criteria (Faithfulness, Conciseness, Readability, Aesthetics, Diversity) select the variant they perceive as better, or Tie when the two are perceived as equivalent.
The first four criteria are rated on 10 page-level comparisons; Diversity is additionally rated on 10 grouped comparisons in which each side shows three landing pages generated for the same paper.
The left/right slot assignment is randomized per participant and per item.
We collected responses from $N{=}23$ participants who completed all 50 questions (40 metric + 10 diversity), yielding 230 responses per criterion.
We report the proportion of Ours / Baseline / Tie selections per criterion in Figure~\ref{fig:userstudy}.

Participants preferred Ours across all criteria (56--65\% vs.\ 21--30\%).
A mixed-effects logistic regression confirms that this preference is statistically significant on all five criteria (OR 2.07--4.05, all $p < 10^{-5}$, Holm-corrected).
The gap is largest on Diversity and Aesthetics, where Baseline's repeated layouts and weaker visual presentation visibly hurt perceived quality, and smallest on Conciseness, where both methods build messaging from the same target specification.
These pairwise human preferences are consistent with the LLM-as-a-Judge and automatic diversity scores in Table~\ref{tab:main-evaluation-results}, supporting the validity of our automatic evaluation.

\subsection{Ablation Study}
We conduct an ablation study to examine the contributions of reference filtering, LandingBench, RAG-based reference retrieval, and individual pipeline phases.
All ablation conditions use gemini-2.5-pro. 
We use seven target specifications, generate three outputs per specification through repeated runs, and evaluate 21 samples per condition, resulting in 126 generated samples across six conditions.

\begin{table}[t]
    \centering
    \small
    \setlength{\tabcolsep}{3.2pt}
    \begin{tabular*}{\columnwidth}{@{\extracolsep{\fill}}lccccc}
        \toprule
        Method & Div. & Faith. & Conc. & Read. & Aesth. \\
        \midrule
        Hard filter only & 0.29 & 4.00 & 4.43 & 4.00 & 3.86 \\
        w/o LandingBench & 0.25 & 4.24 & 4.48 & 4.00 & 3.86 \\
        w/o RAG & 0.29 & 4.10 & 4.38 & 4.00 & 3.86 \\
        w/o Profiling & 0.32 & 4.24 & 4.52 & 4.00 & 3.95 \\
        w/o Polishing & 0.24 & 1.14 & 1.76 & 3.38 & 2.24 \\
        Ours & \textbf{0.38} & \textbf{4.33} & \textbf{4.71} & \textbf{4.05} & \textbf{4.00} \\
        \bottomrule
    \end{tabular*}
    \caption{Ablation results for reference filtering, LandingBench availability, reference retrieval, and pipeline phases. Scores are rounded to two decimal places.}
    \label{tab:ablation-results}
\end{table}

Table~\ref{tab:ablation-results} shows that the full pipeline achieves the highest score on every metric.
\textit{Hard filter only} retains references passing the hard filter during LandingBench construction but omits soft filtering and human verification.
\textit{w/o LandingBench} replaces LandingBench with 438 randomly selected Common Crawl samples as its RAG reference pool while preserving the retrieval mechanism and pool size, thereby isolating LandingBench's contribution.
\textit{w/o RAG} removes LandingBench-based retrieval.
\textit{w/o Profiling} and \textit{w/o Polishing} omit their respective phases, with the latter using the generated page draft as the final output.

\textit{w/o LandingBench} reduced performance across all five metrics, with the strongest relative effect on Diversity.
This indicates that LandingBench contributes to structural variation, content quality, and presentation quality beyond retrieval alone.
Overall, the ablations isolate the contributions of reference filtering, pool composition, retrieval, Profiling, and Polishing to the full pipeline.

\section{Conclusion}
\label{sec:conclusion}

We studied target-grounded, reference-guided landing-page generation. We introduced LandingBench, a reference-profile dataset that abstracts real-world landing pages into reusable structural, rhetorical, and visual representations, and proposed LandingAgent, a three-phase agentic framework that separates Profiling, Wireframing, and Polishing. This decomposition enables the system to structure the target specification, adapt reference patterns for global page planning, and refine the generated page for message quality, visual presentation, and functional consistency.

Experiments show that LandingAgent improves over direct prompting in faithfulness, conciseness, readability, aesthetics, and diversity. These results suggest that abstract reference profiles from real landing pages support target-grounded, diverse generation. Future work should evaluate generated pages in deployed settings and study how perceived page quality relates to real conversion outcomes.


\clearpage

\section*{Limitations}
\label{sec:limitations}

Our work has several limitations. First, because the ultimate goal of a landing page is to induce real user actions, our evaluation provides proxy evidence rather than direct evidence of conversion effectiveness. We assess generated pages through LLM-as-a-Judge, human evaluation, and automatic diversity metrics, but we do not deploy them in real-world environments or measure behavioral outcomes such as click-through or conversion rates. Second, LandingBench serves as a retrieval-based reference pool, and its usefulness depends on the coverage of the collected pages. When a target belongs to an underrepresented industry, language, region, or design convention, the Retriever may only provide approximate references rather than well-matched ones. Third, the main evaluation uses 30 target specifications, and the ablation study uses seven. Although the targets span multiple industry categories and each ablation condition includes three independent generations per target, these repeated runs do not substitute for broader target-level coverage. Because evaluating every ablation condition requires repeated execution of the multi-stage generation and evaluation pipeline, computational cost limited the scale of this analysis. The ablation results should therefore be interpreted as targeted evidence about component contributions rather than exhaustive evidence of generalization. Future work should evaluate generated pages in deployed settings with real user behavior, expand LandingBench toward broader multilingual, domain-specific, and style-diverse reference coverage, and assess the system on larger target sets with stronger statistical analyses.

\bibliography{main}
\clearpage
\appendix
\appendix

\renewcommand{\theequation}{A.\arabic{equation}}
\section{LandingBench Construction Details}
\label{app:landingbench-details}

This appendix provides additional details on the LandingBench construction process, including candidate collection, the three-stage filtering pipeline, reference-profile annotation, and dataset statistics.

\subsection{Candidate Collection}

Landing-page candidates are collected from Common Crawl, a large-scale public web archive~\cite{commoncrawl}.
We crawl approximately 1{,}000 domain-level landing pages drawn from a random sample of the archive. For each page we store the final URL, domain, page title (if available), rendered full-page screenshot, extracted visible text, and basic fetch metadata. Pages that fail to render or return an HTTP error during crawling are recorded with a failure status and excluded from all downstream processing.

\subsection{Filtering Pipeline Overview}

The pipeline applies three sequential stages to narrow the initial 1{,}000 candidates to the final LandingBench set.
The hard filter is applied first and combines rule-based checks, VLM eligibility judgment, and near-duplicate removal into binary pass/fail criteria that remove ineligible, unsafe, and duplicate pages; the soft filter then scores each surviving page on a four-dimensional quality rubric; and a final human verification stage produces the final set.
The main text and Figure~\ref{fig:dataset-pipeline} refer to the hard filter as \emph{rule-based filtering} because its decisions are dominated by deterministic rule and label checks, and to the soft filter as \emph{LLM-based scoring} because it produces continuous LLM-assigned quality scores rather than pass/fail labels.
Table~\ref{tab:pipeline-funnel} summarizes the candidate counts at each stage.

\begin{table}[h]
\centering
\small
\begin{tabular}{lrr}
\toprule
\textbf{Stage} & \textbf{Output} & \textbf{Filtered} \\
\midrule
Initial crawl                                          & 1{,}000 & --- \\
After Hard Filter                                      &   938   &  62 \\
After Soft Filter                                      &   636   & 302 \\
After Human Filter                                     &   438   & 198 \\
\bottomrule
\end{tabular}
\caption{Candidate counts at each stage of the LandingBench filtering pipeline. The final 438 pages are those that passed all three filtering stages and have valid extracted input images and annotation files.}
\label{tab:pipeline-funnel}
\end{table}

\subsection{Hard Filter}

The hard filter is applied first and removes 62 pages from the initial 1{,}000 candidates, leaving 938 pages for soft filtering.
It consists of three sequential stages: rule-based filtering, VLM eligibility judgment, and near-duplicate removal.

\subsubsection{Stage 1: Rule-Based Filtering}

The first stage applies deterministic rules to remove pages with clear non-landing-page signals, safety issues, or insufficient content.
Table~\ref{tab:stage1-rules} summarizes the exclusion categories and criteria.

\begin{table}[h]
\centering
\small
\begin{tabularx}{\columnwidth}{>{\raggedright\arraybackslash}p{0.85in}L}
\toprule
\textbf{Category} & \textbf{Criteria} \\
\midrule
Adult or unsafe content
    & Slug-level keywords (\texttt{porn}, \texttt{sex}, \texttt{xxx}, \texttt{nude}, \texttt{adult}, etc.); multilingual visible-text keywords (English, Arabic, Hindi); gambling and spam signals \\
\midrule
Inaccessibility
    & HTTP 4xx/5xx fetch status; screenshot file size below 30~KB; visible-text keywords indicating error pages, login walls, bot verification, geographic restrictions, or JavaScript requirements \\
\midrule
Insufficient content
    & Visible-text word count below 80; absence of all structural signals (H1, CTA, hero) simultaneously \\
\midrule
Page-type mismatch
    & URL path patterns matching \texttt{/blog/}, \texttt{/news/}, \texttt{/docs/}, \texttt{/jobs/}, \texttt{/wiki/}, \texttt{/help/}, \texttt{/p/\textbackslash d} \\
\bottomrule
\end{tabularx}
\caption{Rule-based exclusion criteria applied in Hard Filter Stage 1.}
\label{tab:stage1-rules}
\end{table}

\subsubsection{Stage 2: VLM Eligibility Judgment}
\label{app:vlm-eligibility}

The second stage applies \texttt{Qwen3.6\_35B\_A3B}~\cite{bai2025qwen3} to classify each remaining page on six eligibility dimensions from a single vision--language model request.
The model receives the page's $1024 \times 1024$ screenshot and the first 600 characters of the visible text.
Screenshots are resized so that the width equals 1024 pixels, then either cropped to the top 1024 pixels (tall pages) or padded with white at the bottom (short pages).

The six eligibility dimensions are:

\begin{itemize}
    \item \textbf{accessibility}: whether the page is accessible and not blocked by errors, login requirements, or verification screens;
    \item \textbf{page\_type}: whether the page is a landing page rather than an article, documentation page, product-detail page, or listing page;
    \item \textbf{content\_depth}: whether the page contains sufficient information for reference annotation;
    \item \textbf{render\_quality}: whether the rendered screenshot is complete and visually interpretable;
    \item \textbf{content\_quality}: whether the visible content appears legitimate and suitable for research use;
    \item \textbf{annotability}: whether section boundaries and page structure can be identified with sufficient reliability.
\end{itemize}

The classification prompt is:

\begin{mdframed}[style=promptbox]
\begin{small}
\begin{Verbatim}[breaklines=true,breakanywhere=true]
[PAGE TEXT (first 600 chars)]
{visible_text[:600]}

Evaluate this landing page screenshot on 6 criteria.
Answer with ONLY the labels listed.

1. accessibility
   Labels: accessible | error_page | login_required
           | geo_blocked | bot_challenge
2. page_type
   Labels: landing_page | blog_post | news_article
           | documentation | job_listing
           | product_detail | other
3. content_depth
   Labels: sufficient | minimal | hero_only
4. render_quality
   Labels: normal | minor_issues | broken
5. content_quality
   Labels: legitimate | adult | gambling | spam_phishing
6. annotability
   Labels: annotatable | ambiguous | not_annotatable

Output ONLY valid JSON, nothing else:
{"accessibility":"...","page_type":"...","content_depth":
"...","render_quality":"...","content_quality":"...",
"annotability":"..."}
\end{Verbatim}
\end{small}
\end{mdframed}

A page passes Stage 2 if and only if it satisfies all six of the following conditions:

\begin{itemize}
    \item \texttt{accessibility} $=$ \texttt{accessible}
    \item \texttt{page\_type} $\notin$ \{\texttt{blog\_post}, \texttt{news\_article}, \texttt{documentation}, \texttt{product\_detail}, \texttt{job\_listing}\}
    \item \texttt{content\_depth} $\neq$ \texttt{hero\_only}
    \item \texttt{render\_quality} $\neq$ \texttt{broken}
    \item \texttt{content\_quality} $=$ \texttt{legitimate}
    \item \texttt{annotability} $\neq$ \texttt{not\_annotatable}
\end{itemize}

Model outputs are parsed using a three-step fallback: (1) a JSON block following a \texttt{</think>} tag or in a Markdown code fence; (2) a bare JSON object anywhere in the response; (3) per-label keyword extraction from the full response text.

\subsubsection{Stage 3: Near-Duplicate Removal}

The third stage removes near-duplicate pages using complementary visual and textual signals.

\paragraph{Visual deduplication.}
We compute a 64-bit perceptual hash (pHash) for each screenshot.
Page pairs whose pHash Hamming distance is at most 10 are flagged as visually near-duplicate.

\paragraph{Textual deduplication.}
We compute TF-IDF vectors over the visible-text content of each page.
Page pairs with cosine similarity of at least 0.92 are flagged as textually near-duplicate.

\paragraph{Pair resolution.}
For each detected near-duplicate pair, we retain the page with the higher soft-filter total score (sum of IC, BSC, LR, VPQ) and discard the other.

\subsection{Soft Filter: LLM-Based Quality Scoring}
\label{app:subsec:soft_filter}

The soft filter is applied to the 938 hard-filter survivors and removes 302 additional pages, leaving 636 pages in the LandingBench set.

It applies \texttt{Qwen3.6\_35B\_A3B} to score each page on four quality dimensions using the page's rendered screenshot and extracted visible text.

A compact brand brief is constructed from each page's extracted brand metadata and prepended to the visible text. The model receives the full-page screenshot and complete visible text of each candidate page and assigns a score from 1 to 5 for each dimension in Table~\ref{tab:soft-filtering}. A page is passed to human verification only if it receives a score of 4 or higher on every dimension.

The scoring prompt is:

\begin{mdframed}[style=promptbox]
\begin{small}
\begin{Verbatim}[breaklines=true,breakanywhere=true]
[BRAND BRIEF]
{brief}

[PAGE TEXT]
{vtext}

Score 1-5 (1=worst, 5=best):
ic=Information Completeness,
bsc=Brand-Style Consistency,
lr=Layout Readability,
vpq=Visual Presentation Quality

Output ONLY valid JSON, nothing else:
{"ic":N,"bsc":N,"lr":N,"vpq":N}
\end{Verbatim}
\end{small}
\end{mdframed}

\begin{table*}[h]
\centering
\small
\begin{tabularx}{\textwidth}{lLc}
\toprule
\textbf{Dimension} & \textbf{Description} & \textbf{Score} \\
\midrule
Information Completeness
& Product/service information and the value proposition are sufficiently presented
& 1--5 \\
Brand-Style Consistency
& Tone and visual style are coherent across the page
& 1--5 \\
Layout Readability
& Section boundaries and information flow are clear
& 1--5 \\
Visual Presentation Quality
& Overall visual completeness and polish are sufficient
& 1--5 \\
\bottomrule
\end{tabularx}
\caption{Soft filtering rubric for estimating reference suitability.}
\label{tab:soft-filtering}
\end{table*}

\subsection{Human Verification Protocol}

After hard filtering and soft filtering, two human reviewers independently verify each candidate page. Table~\ref{tab:human-verification} summarizes the verification protocol.

\begin{table*}[h]
\centering
\small
\begin{tabularx}{\textwidth}{lL}
\toprule
\textbf{Verification item} & \textbf{Description} \\
\midrule
Inputs to reviewers
& 636 original webpages; full-page screenshots; complete visible text \\
Annotators            & 2 independent reviewers \\
Rejection rule        & Rejected if either annotator assigns \textsc{Reject} \\
\midrule
\multicolumn{2}{l}{\textit{Verification criteria}} \\
\midrule
Landing-page validity
    & Is the page a genuine landing page? Articles, portals, blogs, and documentation are rejected. \\
Annotability
    & Can the page be reliably segmented into sections with clearly identifiable boundaries? \\
Render quality
    & Does the page render correctly without critical layout failures? \\
CTA presence
    & Does the page contain at least one identifiable call-to-action? \\
Content safety
    & Is the content appropriate and safe for research use? \\
\midrule
Accepted              & 438 pages (68.9\%) \\
Rejected              & 198 pages (31.1\%) \\
\bottomrule
\end{tabularx}
\caption{Human verification protocol for final LandingBench candidate selection.}
\label{tab:human-verification}
\end{table*}

\subsection{Screening Agreement and Rejection Analysis}

Across all 636 candidates, the two reviewers' binary decisions yielded an observed agreement of $P_o=0.821$ and Cohen's $\kappa=0.48$ (95\% CI: $[0.40, 0.56]$), indicating moderate agreement.
Disagreements were concentrated among ambiguous B2B SaaS and app/tool pages.
Because acceptance required both reviewers to assign \textsc{Accept}, the final set represents the conservative intersection of their decisions.

Annotators are instructed to reject pages that are primarily articles, blog posts, documentation pages, support pages, job postings, product-detail pages, catalog listings, login pages, or other non-landing-page interfaces. Pages are also rejected if their section boundaries are too ambiguous for reliable annotation.

\begin{table}[h]
\centering
\small
\begin{tabular}{lrr}
\toprule
\textbf{Exclusion type} & \textbf{Count} & \textbf{\%} \\
\midrule
News or editorial              & 77 & 38.9 \\
Ad-tech or infrastructure      & 27 & 13.6 \\
E-commerce                     & 16 &  8.1 \\
Other exclusion categories     & 78 & 39.4 \\
\midrule
\textbf{Total}                 & \textbf{198} & \textbf{100.0} \\
\bottomrule
\end{tabular}
\caption{Exclusion-type distribution of the 198 candidates rejected during human verification.}
\label{tab:human-rejection-types}
\end{table}

Overall, 96\% of rejected pages corresponded to identifiable non-landing-page types, consistent with the exclusion rationale applied during human verification.
The remaining cases (approximately 4\%) were borderline, primarily B2B SaaS and app/tool pages that more closely resembled product dashboards or functional application interfaces than public-facing landing pages.

\subsection{Reference Profile Annotation}

Each accepted page is converted into a reference profile. Table~\ref{tab:reference-profile-components} summarizes the components currently specified in the main manuscript and additional components that require final taxonomy design.

\begin{table*}[h]
\centering
\small
\begin{tabularx}{\textwidth}{lL}
\toprule
\textbf{Component} & \textbf{Description} \\
\midrule
\multicolumn{2}{l}{\textit{Page structure}} \\
\midrule
Rendered reference screenshot
    & Full-page screenshot used as visual context during generation \\
Structured section sequence
    & Ordered list of sections, each annotated with a role label (\texttt{kind}), purpose, headline, key copy excerpts, visual description, and component list \\
Section-role labels
    & Per-section label from a 13-class soft enum: features, other, footer, hero, cta band, testimonials, product tour, metrics, logos strip, use cases, pricing, faq, integrations \\
CTA structure
    & Primary and secondary CTA copy with placement-pattern description \\
Page-structure axes tuple
    & Structured retrieval axes: message length, information density, scroll depth, CTA structure, asset dependency \\
\midrule
\multicolumn{2}{l}{\textit{Visual identity}} \\
\midrule
Color palette descriptor
    & Primary, neutral, and accent hex values with semantic-usage description \\
Typography descriptor
    & Headline and body style descriptions with scale notes; font-family names are not extracted \\
Imagery \& iconography style
    & Free-text description of imagery approach, iconography treatment, motion hints, and overall mood \\
\midrule
\multicolumn{2}{l}{\textit{Brand metadata}} \\
\midrule
Brand archetype
    & Primary and secondary Jungian archetype labels; 12 classes: Sage, Creator, Ruler, Caregiver, Hero, Everyman, Magician, Outlaw, Explorer, Jester, Innocent, Liberator \\
Company stage
    & Maturity stage inferred from page signals: Late-stage, Public, Growth, Early-stage, Unknown \\
Brand polish level
    & Perceived production quality: High-polish, Scrappy/Minimal, Luxury/Premium, Unknown \\
Value proposition \& audience
    & One-line pitch, target audience segments, and key value propositions extracted from the page \\
Social proof signals
    & Customer logo list, claimed metrics, and testimonial excerpts \\
Tone \& voice attributes
    & Tone-of-voice description and voice-attribute list extracted from copy \\
\bottomrule
\end{tabularx}
\caption{Reference profile components extracted for each LandingBench page, organized by category.}
\label{tab:reference-profile-components}
\end{table*}

\subsection{Annotation Guidelines}

Reference profiles are constructed automatically using vision--language models (VLMs) rather than manual human annotation.
Each page is processed through two extraction pipelines that operate on the same inputs: a full-page screenshot resized to $1024 \times 1024$ pixels and the extracted visible text.
Screenshots are resized so that the width equals 1024 pixels, then either cropped to the top 1024 pixels for tall pages or padded with white at the bottom for short pages.

\paragraph{Website structure extraction.}
\texttt{Claude Sonnet 4.6}~\cite{anthropic2026sonnet46} is prompted to analyze the page screenshot and visible text and produce a structured JSON annotation covering:

section sequence (\texttt{sections\_in\_order}, each with a \texttt{kind} label, purpose, headline, key copy excerpts, visual description, and component list);
visual identity (color palette with hex values and semantic-usage description, typography style notes, imagery and iconography characterization, motion hints, and overall mood);
layout (grid pattern, navigation items and primary CTA, hero content);
CTA structure (primary and secondary CTA copy with placement-pattern description);
company metadata (name, URL, one-line pitch, target audience, value propositions); and
social proof signals (customer logo list, claimed metrics, testimonial excerpts).
The output is stored as \texttt{website\_extraction.json}.

\paragraph{Brand information extraction.}

\texttt{Claude Sonnet 4.6}~\cite{anthropic2026sonnet46} is prompted with the same screenshot and visible text and produces a deeper brand analysis stored as \texttt{brand\_extraction.json}.

The extracted fields include:
founding thesis (origin hypothesis, worldview, mission, vision, core values);
brand strategy (primary and secondary Jungian archetype, positioning statement, brand promise, personality traits);
market and audience (market category, ideal customer profile, buyer personas, jobs-to-be-done);
brand expression (voice attributes, tone modulation by section, narrative arc, vocabulary signatures);
visual brand rationale (color strategy, typography and imagery rationale, layout density, logo strategy);
maturity signals (company stage, brand polish level); and
an inverse pipeline input block encoding the page as a structured generation brief.

\paragraph{Output validation.}
Both extraction outputs are validated against a JSON schema.
Responses that fail to parse as valid JSON are repaired using the \texttt{json\_repair} library before schema validation.
The validated outputs from both pipelines constitute each accepted page's reference profile.

\subsection{Dataset Statistics}

After the two-stage automated pipeline and one-stage human filtering, LandingBench contains 438 verified landing pages. Table~\ref{tab:dataset-stats-market} shows the market-category distribution.
Categories are normalized from free-text market descriptions generated by the VLM into 14 broad industry labels.

\begin{table}[h]
\centering
\small
\begin{tabular}{lrr}
\toprule
\textbf{Market category} & \textbf{Count} & \textbf{\%} \\
\midrule
DevTools / Infra          &  82 & 18.7 \\
Media / Content           &  80 & 18.3 \\
Security                  &  61 & 13.9 \\
B2B SaaS                  &  49 & 11.2 \\
Marketing / Analytics     &  29 &  6.6 \\
Other                     &  27 &  6.2 \\
AdTech                    &  26 &  5.9 \\
Education                 &  25 &  5.7 \\
Finance / Fintech         &  18 &  4.1 \\
E-commerce                &  15 &  3.4 \\
Social / Community        &  13 &  3.0 \\
Gaming                    &   9 &  2.1 \\
Travel                    &   2 &  0.5 \\
Health / Bio              &   2 &  0.5 \\
\bottomrule
\end{tabular}
\caption{Market-category distribution of LandingBench pages.}
\label{tab:dataset-stats-market}
\end{table}

\begin{table*}[h]
\centering
\small
\begin{tabular}{llrr}
\toprule
\textbf{Axis} & \textbf{Category} & \textbf{Count} & \textbf{\%} \\
\midrule
\multirow{8}{*}{Brand archetype}
    & Sage            & 107 & 24.4 \\
    & Ruler           &  61 & 13.9 \\
    & Creator         &  59 & 13.5 \\
    & Caregiver       &  50 & 11.4 \\
    & Hero            &  42 &  9.6 \\
    & Magician        &  27 &  6.2 \\
    & Everyman        &  27 &  6.2 \\
    & Other / Unknown &  65 & 14.8 \\
\midrule
\multirow{5}{*}{Company stage}
    & Late-stage  & 206 & 47.0 \\
    & Public      & 112 & 25.6 \\
    & Growth      &  89 & 20.3 \\
    & Unknown     &  29 &  6.6 \\
    & Early-stage &   2 &  0.5 \\
\midrule
\multirow{4}{*}{Brand polish level}
    & High-polish       & 397 & 90.6 \\
    & Unknown           &  27 &  6.2 \\
    & Scrappy / Minimal &   8 &  1.8 \\
    & Luxury / Premium  &   6 &  1.4 \\
\midrule
\multirow{6}{*}{Primary color hue}
    & Blue / Purple          & 211 & 48.2 \\
    & Red / Orange           & 121 & 27.6 \\
    & Black / Dark           &  33 &  7.5 \\
    & Green                  &  30 &  6.8 \\
    & Red / Pink             &  24 &  5.5 \\
    & Gray / White / Unknown &  19 &  4.3 \\
\bottomrule
\end{tabular}
\caption{Distributions of brand archetype, company stage, brand polish level, and primary color hue across the 438 LandingBench pages.
Brand metadata is extracted using Claude Sonnet 4.6 from each page's screenshot and visible text.
Archetype and stage categories are normalized from free-text model output.
Color hue is derived from the primary palette hex value.}
    \label{tab:dataset-stats-distributions}
\end{table*}

Table~\ref{tab:dataset-stats-distributions} shows the distributions of brand archetype, company stage, brand polish level, and primary color hue.

\renewcommand{\theequation}{B.\arabic{equation}}
\section{LandingAgent Implementation Details}
\label{app:landingagent-details}

This appendix documents the prompt configuration of each agent in LandingAgent. Table~\ref{tab:agent-prompt-summary} summarizes the inputs and output schemas of all eight agents.
Listings~\ref{lst:profiler_prompt}--\ref{lst:functional_critic_prompt} show the prompt templates of all eight agents in execution order: \textsc{Profiler}, \textsc{Retriever}, \textsc{WireframeBuilder}, \textsc{WireframeCritic}, \textsc{Polisher}, \textsc{AestheticCritic}, \textsc{MessageCritic}, and \textsc{FunctionalCritic}.
All eight prompts share a common structural pattern (\textsc{Role} / \textsc{Task} / \textsc{Input} / \textsc{Output schema} / per-field guidance). 

\begin{table*}[!t]
\centering
\small
\begin{tabularx}{\textwidth}{llLL}
\toprule
\textbf{Agent} & \textbf{Phase} & \textbf{Inputs} & \textbf{Output} \\
\midrule
Profiler
& Profiling
& Target specification, optional target images
& Structured Page Brief (JSON): value prop, audience, tone, must-have sections, 11 page-structure axes, 5 visual axes \\
Retriever
& Wireframing
& Structured Page Brief, LandingBench profiles, optional restart hint
& Selected reference profile (id, rationale) \\
WireframeBuilder
& Wireframing
& Structured Page Brief, reference profile
& Page draft (HTML) \\
WireframeCritic
& Wireframing
& Rendered draft, page draft, Structured Page Brief, reference profile
& \textsc{Proceed}/\textsc{Restart} decision with diagnostic hint \\
Polisher
& Polishing
& Page draft, Structured Page Brief, style guide, critic feedback
& Refined page implementation \\
AestheticCritic
& Polishing
& Rendered page, reference profile, Structured Page Brief
& Score (1--5), approval decision, visual feedback, optional patch \\
MessageCritic
& Polishing
& Page copy, Structured Page Brief, generated implementation
& Score (1--5), approval decision, message feedback, optional patch \\
FunctionalCritic
& Polishing
& Generated implementation, rendered page metadata
& Score (1--5), approval decision, functional feedback, optional patch \\
\bottomrule
\end{tabularx}
\caption{Agent prompt summary. Each agent's prompt follows a common \textsc{Role} / \textsc{Task} / \textsc{Input} / \textsc{Output schema} structure.}
\label{tab:agent-prompt-summary}
\end{table*}

\begin{figure*}[!p]
\begin{lstlisting}[
  basicstyle=\small\ttfamily,
  breaklines=true,
  frame=single,
  caption={Prompt for Profiler},
  columns=fullflexible,
  label={lst:profiler_prompt},
  keepspaces=true,
  literate={-}{-}1
]
## ROLE
You are a senior brand strategist preparing a concise creative brief for an AI landing-page design team.

## TASK
Analyze the supplied target specification (free-form text describing the product, audience, tone, palette, constraints) and optional target images (logos, product UI, team photos, illustrations). Extract only brand facts and plausible design implications supported by the input. Never invent product claims, metrics, customers, awards, integrations, or pricing details absent from the input. When the input does not support an inference: scalar fields -> null; list fields -> []. Infer two matching-axis blocks: page_structure_axes (11 structural axes) and visual_axes (5 visual/brand axes) used for matching against reference websites. Use enum values exactly.

## INPUT DATA
- Raw target specification as text.
- Optional target images (logos, product UI screenshots, team photos, illustrations, etc.).

## OUTPUT  (strict JSON, no markdown)
{
  "company_name":      <verbatim brand/product name, or null>,
  "target_audience":   <string or null>,
  "value_prop":        <string or null>,
  "tone_descriptors":  [3-6 tone words],
  "visual_identity_hints": { "palette_hints": [...], "typography_hints": [...], "imagery_style": <string or null> },
  "must_have_sections": [<sections derived from spec>],
  "brand_keywords":     [...], "industry": <string or null>, "brand_description": <1-3 sentences>,
  "funnel_intent":      <enum: educate-convert | show-convert | trust-convert | ...(continue)... | null>,
  "page_structure_axes": {
      "message_count": <integer or null>, "info_density": <enum: sparse | medium | dense | null>,
      "scroll_depth":  <enum: short | medium | long | null>,
      ...(continue with 8 more axes: section_kinds, proof_types, cta_structure, cta_density, audience_structure, interactivity, nav_complexity, hero_pattern)...
  },
  "visual_axes": {
      "palette_mode": <enum: dark-dominant | light-dominant | brand-saturated | monochrome-accent | null>,
      "palette_size": <enum: minimal | standard | rich | null>,
      ...(continue with 3 more axes: typography_character, mood_tags, maturity_polish)...
  }
}

## PAGE_STRUCTURE_AXES GUIDANCE
Short rule per axis; e.g., scroll_depth: short <=4 sections, medium 5-7, long 8+. ...(continue)...

## VISUAL_AXES GUIDANCE
Short rule per axis; e.g., palette_size: minimal <=2 brand colors; standard 3-4; rich 5+. ...(continue)...

## SECTIONS GUIDANCE
- Derive each entry only from content present in the spec; honor explicit prohibitions.
- Keep must_have_sections consistent with page_structure_axes (omit "proof" when proof_types is []; omit "cta" when funnel_intent is awareness-only).
- Valid entries: hero, features, how_it_works, demo, proof, pricing, faq, cta, footer.

## USER MESSAGE (filled at runtime)
Analyze this target specification for a landing-page.
Target specification: {spec_body}
Target image paths: {target_image_paths}
[Target image 1: <path>] <image_part_1> ...
\end{lstlisting}
\end{figure*}

\begin{figure*}[!p]
\begin{lstlisting}[
  basicstyle=\small\ttfamily,
  breaklines=true,
  frame=single,
  caption={Prompt for Retriever},
  columns=fullflexible,
  label={lst:retriever_prompt},
  keepspaces=true,
  literate={-}{-}1
]
## ROLE
You are the structural + visual retrieval agent for an AI landing-page design system. The reference you select seeds both the prototype scaffold and the final polished page, so structural shape AND visual character must hold simultaneously.

## TASK
Score each candidate reference against the brief's 16 matching axes (11 page_structure_axes $+$ 5 visual_axes) using the weighted criteria below, apply hard-reject rules, and return the single best candidate id. Match on page-shape and visual character - NOT industry or tone words. If iterative critic feedback is supplied, re-rank to address the prior failure while preserving hard-reject constraints.

## INPUT DATA
- Brief page_structure_axes (11) + visual_axes (5) + full brand brief for context.
- judge_hints: critic feedback placeholder; (none) on iteration 0, free-text on iteration >= 1.
- Previously Selected (rejected): ref ids the critic rejected; empty on iteration 0.
- Candidate pool: each item has id, page_structure_axes, visual_axes.

## OUTPUT  (strict JSON, no markdown)
{
  "top_k_reference_ids": [<single best candidate id>]
}

## MATCHING - 16 axes, two tiers
Tier 1 - Structural carrier (page-shape; primary):
- 3.0: section_kinds, message_count, scroll_depth
- 2.5: proof_types, cta_structure
- 2.0: audience_structure, hero_pattern
- 1.5: cta_density, interactivity, nav_complexity, info_density
Tier 2 - Visual/brand carrier (look-and-feel; polish):
- 1.5: palette_mode, mood_tags
- 1.0: typography_character, maturity_polish, palette_size

## HARD-REJECT RULES (override aggregate score)
- proof_types: reject if candidate's core hero or primary structure depends on a proof type the brief lacks (e.g., stat-band hero, brief has no stats). Extras in optional sections do NOT trigger reject.
- interactivity: reject if ref is embedded-demo but brief is static.

## NULL vs EMPTY  (treated differently)
- Brief axis null = unknown -> skip from weighted sum; cannot trigger hard-reject.
- Candidate axis null = weak evidence -> cap that axis at OK (never strong, never mismatch).
- Empty list [] is axis-specific: proof_types: [] = explicitly no proof (real value; can trigger hard-reject); section_kinds: [] and mood_tags: [] = treat as unknown.

## CRITIC LOOP
If judge_hints == (none) AND Previously Selected (rejected) empty -> iteration 0; score on axes only. Otherwise re-rank to address the critic's complaint; keep a previously-rejected id only if hints explicitly say it still fits on the non-problem dimension. Hard-reject rules are inviolable regardless of hints.

## USER MESSAGE (filled at runtime)
Select the single best reference whose page_structure_axes + visual_axes fit the brief.
Brief page_structure_axes (Tier 1): {brief_page_structure_axes}
Brief visual_axes (Tier 2):        {brief_visual_axes}
Brief (full, context only):        {brand_brief}
Judge Hints:                       {judge_hints | "(none)"}
Previously Selected (rejected):    {previous_reference_ids | "[]"}
Candidate Pool (id + page_structure_axes + visual_axes):
  [Candidate i: <id>, <page_structure_axes>, <visual_axes>] ...
Return strictly valid JSON with top_k_reference_ids (a single-element list).
\end{lstlisting}
\end{figure*}

\begin{figure*}[!p]
\begin{lstlisting}[
  basicstyle=\small\ttfamily,
  breaklines=true,
  frame=single,
  caption={Prompt for WireframeBuilder},
  columns=fullflexible,
  label={lst:wireframe_builder_prompt},
  keepspaces=true,
  literate={-}{-}1
]
## ROLE
You are a senior landing-page WireframeBuilder creating a WIREFRAME -- a structural shell with generic placeholder content. A separate polisher pass fills in real brand copy and brand images afterwards using the full brief. Do not try to communicate the brand in this stage.

## TASK
Generate one standalone static HTML document capturing the page's STRUCTURE only. Use Tailwind via CDN. Use semantic sections matching the brief's must_have_sections. Apply skeleton-level Tailwind styling that reflects visual_identity_hints (palette tone, typography character) -- just enough so the page reads as the right design family. ALL copy must be generic placeholder text; ALL images must be empty visual slots. Do not invent product specifics; the polisher will replace placeholders with real content.

## INPUT DATA  (narrowed -- structure-only signals)
You receive: must_have_sections; page_structure_axes; visual_axes; visual_identity_hints; hydrated top_k_references (compact + screenshot). ...(continue: per-field detail -- which axes, what's stripped, etc.)...
You do NOT receive (these go to the polisher): target_audience, value_prop, tone_descriptors, brand_keywords, industry, brand_description, full brand_brief, brand image paths.

## OUTPUT  (strict JSON, no markdown)
{ "html": "<!doctype html><html>...</html>" }
(No assets field -- the wireframe references no real images.)

## REFERENCE MIRRORING  (your default -- deviate only if the brief explicitly contradicts)
Section inclusion is determined by the structural brief first: include every section required by must_have_sections plus structural slots implied by page_structure_axes / visual_axes. ...(continue)...
Mirror these from the reference:
- hero_pattern (split / centered / full-bleed-visual / minimal-text / asymmetric) -- match exactly.
- sections_structured order.
...(continue)...
Palette and typography come from visual_identity_hints, NOT the reference screenshot.

## REJECT THESE OUTPUTS
- Generic Tailwind centered hero with 2 buttons when the reference shows split layout.
...(continue)...

## PLACEHOLDER CONTENT RULES

### Text placeholders -- generic boilerplate ONLY
Every text element gets generic, category-neutral content. Plain readable English. NO lorem ipsum, NO brand-specific copy from input. ...(continue)...

### Image slots -- empty visual placeholders
Every image region is an empty styled div. NEVER use <img> tags. NEVER reference assets/... paths. NEVER use placehold.co. Pattern:
<div class="aspect-video bg-neutral-100 border border-dashed rounded-lg ..." role="img" aria-label="Image placeholder -- hero visual"><span>[ Image: hero visual ]</span></div>
...(continue)...

## SECTION STRUCTURE -- WIREFRAME-LEVEL ONLY
- Every must_have_sections entry -> one semantic <section> (or <header> / <footer>).
- Navigation: include unless must_have_sections or page_structure_axes explicitly excludes; match nav_complexity.
...(continue)...

## USER MESSAGE (filled at runtime)
Build the landing-page DRAFT (structural shell with generic placeholders only -- no brand copy, no real images). Ensure the draft has a structural slot for EVERY section the brief requires -- do not rely solely on must_have_sections.
Structural brief (subset of full brief -- content fields withheld):
{ "must_have_sections": {must_have_sections}, "page_structure_axes": {page_structure_axes}, "visual_axes": {visual_axes}, "visual_identity_hints": {visual_identity_hints} }
References: {top_k_references_compact}
[Reference screenshot <id_1>] <image_part_1>
[Reference screenshot <id_2>] <image_part_2> ...
\end{lstlisting}
\end{figure*}

\begin{figure*}[!p]
\begin{lstlisting}[
  basicstyle=\small\ttfamily,
  breaklines=true,
  frame=single,
  caption={Prompt for WireframeCritic},
  columns=fullflexible,
  label={lst:wireframe_critic_prompt},
  keepspaces=true,
  literate={-}{-}1
]
## ROLE
You are a wireframe critic -- a pragmatic creative director deciding whether a landing-page DRAFT should continue to refinement or restart retrieval.

## TASK
You receive the rendered DRAFT screenshot and the raw HTML. The DRAFT is a structural shell -- placeholder copy and empty image slots are EXPECTED. Critique only the STRUCTURE (section flow, layout patterns, density, hero pattern, CTA structure) against the brief and reference. Do not restart for content-level concerns (those go to the polisher). If the screenshot is unavailable, critique from HTML at lower confidence.

## INPUT DATA
- Brand brief JSON.
- Hydrated top_k_references.
- Raw DRAFT HTML (structural shell with placeholders).
- Rendered DRAFT screenshot when available.

## OUTPUT  (strict JSON, no markdown)
{
  "decision": "proceed" | "restart",
  "reasoning": <evidence-cited explanation citing section ids, hero class names, brief axis values>,
  "hints":     <if restart: which axis/section mismatched and what retrieval criteria to favor next attempt; if proceed: 1-3 concrete refinement targets for the polisher>,
  "axis_checks": {
      "must_have_sections_present": all | partial | n/a,
      "hero_pattern_match":         yes | no | n/a,
      "cta_structure_match":        yes | no | n/a,
      "interactivity_match":        yes | no | n/a,
      "scroll_depth_match":         yes | no | n/a
  },
  "restart_trigger": null | wrong_shape | missing_section | hero_pattern | cta_structure | interactivity | design_meta | domain_clash
}

## RESTART TRIGGERS  (restart=true iff ONE OR MORE holds; else proceed)
1. **Wrong page shape**: the draft's overall section count or scroll depth differs from `brief.page_structure_axes.scroll_depth` by more than one bucket (e.g. brief says `short` but draft has 8+ sections).
2. **Missing must-have section**: `brief.must_have_sections` is non-empty AND at least one named entry has no corresponding `<section>` / `<header>` / `<footer>` in the draft. (List the missing items in `hints`.)
...(continue with 5 more triggers: hero_pattern, cta_structure, interactivity, design_meta, domain_clash)...

Empty brief special case: if `brief.must_have_sections` is empty AND most `page_structure_axes` are null, do NOT invent requirements from the reference; choose `proceed` with low confidence and note "brief under-specified" in `reasoning`.

## USER MESSAGE (filled at runtime)
{verbatim_brief_block}
Critique this landing-page wireframe.

Brand brief:
{brand_brief_json}

References (slim):
{top_k_references_slim_json}   # truncated to 12000 chars

Prototype HTML:
{prototype_html}

Rendered prototype screenshot: {prototype_screenshot_path}  <image_part>
[or, if missing]
Rendered prototype screenshot is missing; critique from HTML only.
\end{lstlisting}
\end{figure*}

\begin{figure*}[!p]
\begin{lstlisting}[
  basicstyle=\small\ttfamily,
  breaklines=true,
  frame=single,
  caption={Prompt for Polisher},
  columns=fullflexible,
  label={lst:polisher_prompt},
  keepspaces=true,
  literate={-}{-}1
]
## ROLE
You are a senior web art director who takes a structural DRAFT and produces the finished landing page. The draft arrives with generic placeholder copy and empty image slots; you fill it with the brand's real content and visual identity using the full brand brief.

## HARD CONSTRAINTS -- THE BRIEF OVERRIDES EVERYTHING
The user message includes the VERBATIM USER BRIEF -- the original brief word-for-word, the single authoritative source of truth. The structured `brand_brief` JSON is a lossy summary; when they disagree, the verbatim brief wins. Any rule phrased as a constraint or prohibition ("must not", "do not", "no <X>", "only", "never", "forbidden", "avoid") is a HARD CONSTRAINT and outranks every other instruction, including critique patches.

## TASK PER ROUND

### Round 0 -- FILL THE DRAFT
Input: `draft_html` + full `brand_brief` + brand image assets + chosen style guide.
1. Replace every generic placeholder copy with brief-derived copy (headlines, CTAs, features, testimonials, stats, pricing, FAQ, footer) reflecting `target_audience`, `value_prop`, `tone_descriptors`, `brand_keywords`, `industry`. Use the supplied `BRAND NAME` string verbatim everywhere; NEVER invent/translate/abbreviate. Keep `must_have_sections` intact EXCEPT fact-data sections with zero brief support ...(continue)...
2. Resolve every empty image slot -- NEVER ship the dashed placeholder. Per slot: (a) brand asset fits -> swap in `<img src="assets/...">`; (b) no asset AND slot is "fact-data" (customer logo, real testimonial face, team photo, press logo, cert mark, real product screenshot) -> remove the ENTIRE section; (c) otherwise -> finished CSS-only treatment from the LIBRARY below.
3. Apply real brand visual identity: swap skeleton Tailwind colors for `visual_identity_hints.palette_hints`; apply `typography_hints`; honor chosen style guide (modern / minimal / playful).
4. Preserve the draft's STRUCTURE: hero pattern, section order, density, CTA structure left intact.

### Round n >= 1 -- APPLY CRITIQUES
Input: previous polisher HTML + aesthetic, message, functional critiques. Apply each critique's `patches` (see APPLYING CRITIQUES). ...(continue)...

## APPLYING CRITIQUES
Critiques arrive as `{"aesthetic": {...}, "message": {...}, "functional": {...}}`, each with `scores`, `issues`, `suggestions`, `patches`, `approved`. Treat `patches` as the primary authoritative actions. For each patch: locate via `target`; apply `op` in {replace, add, remove, style, move, rename}; `style` = Tailwind/inline only.

## VISUAL TREATMENT LIBRARY (for empty image slots; pull colors from `palette_hints`, tune intensity from `tone_descriptors`)

### NO FABRICATED NUMBERS RULE
Treatments may show UI scaffolding but MUST NOT contain invented concrete data. Real metric from brief -> use exact number. No real metric -> use non-numeric placeholder instantly readable as "not real": `00%`, `0,000+`, `00`, `0.0x`, `$0,000`. NEVER plausible-looking values like `12.4%`, `1.2M`, `$4,820`, `99.99%`. Critics know `00%`-style tokens are intentional and will not flag them.

## INPUT DATA
- Current review round; `BRAND NAME` (verbatim); VERBATIM USER BRIEF (authoritative); full brand brief; chosen style guide text; available brand image assets (`assets/...` paths, OPTIONAL); ...(continue)...

## OUTPUT (strict JSON, no markdown)
{ "html": "<!doctype html><html>...</html>", "style_guide": "{guide_name}" }

## USER MESSAGE (filled at runtime)
Current review round: {round_idx}
Selected style guide: {guide_name}
{brand_name_block}
## VERBATIM USER BRIEF -- AUTHORITATIVE SOURCE (overrides the structured brief and every critique)
<<<BEGIN VERBATIM BRIEF>>> {brief_body} <<<END VERBATIM BRIEF>>>
Style guide: {guide_text}
Brand brief (full): {brand_brief_json}
...(continue)...
\end{lstlisting}
\end{figure*}

\begin{figure*}[!p]
\begin{lstlisting}[
  basicstyle=\small\ttfamily,
  breaklines=true,
  frame=single,
  caption={Prompt for AestheticCritic},
  columns=fullflexible,
  label={lst:aesthetic_critic_prompt},
  keepspaces=true,
  literate={-}{-}1
]
## ROLE
You are an aesthetic critic for modern landing pages.

## TASK
Review the rendered screenshot and raw HTML. Score eight dimensions from 1 to 5: five usual visual-quality dimensions (typography_hierarchy, color_harmony, whitespace, alignment, modernness) plus visual_readability, reference_fidelity, and brand_expression. Emit free-text issues/suggestions and structured, region-annotated patches the polisher can apply directly. ...(continue)...

## INPUT DATA
- Review round.
- Raw HTML.
- Rendered screenshot of the page being reviewed when available.
- Reference screenshot (retrieval Top-1) when available -- used ONLY for reference_fidelity.
- Brand brief -- used ONLY to distinguish brief-gap empty slots (acceptable) from polisher omissions (failure), and to score brand_expression. ...(continue)...

## OUTPUT  (strict JSON, no markdown)
{
  "scores": {
    "typography_hierarchy": <1-5>, "color_harmony": <1-5>, "whitespace": <1-5>,
    "alignment": <1-5>, "modernness": <1-5>, "visual_readability": <1-5>,
    "brand_expression": <1-5>, "reference_fidelity": <1-5>
  },
  "issues":      [<short human-readable issue strings>],
  "suggestions": [<short human-readable suggestion strings>],
  "patches": [
    { "target": <CSS-ish selector hint>, "bbox": [x, y, w, h],
      "current_value": <short snippet or null>, "suggested_value": <short snippet>,
      "op": replace | add | remove | style | move,
      "rationale": <one sentence visual reason>, "severity": low | medium | high }
  ]
}

## SCORING DIMENSIONS
- typography_hierarchy, color_harmony, whitespace, alignment, reference fidelity -- usual visual-quality dimensions.
- visual_readability -- scanability and contrast of rendered text (font size at viewport scale, line-height, foreground/background separation, heading-to-body hierarchy). ...(continue)...
- reference_fidelity -- when a reference is supplied, how well the rendered page captures the reference's STRUCTURE (layout pattern, section flow, visual rhythm, density, asset placement). ...(continue)...
- brand_expression -- how strongly the VISUAL design conveys THIS specific brand's personality per the brief. ...(continue)...

## DOWNSTREAM APPROVAL GATE (for reference)
approved is computed deterministically downstream from your scores: rendered-page screenshot must be available AND the six core dimensions (typography_hierarchy, color_harmony, whitespace, alignment, modernness, visual_readability) all >= 4. ...(continue)...

## STRUCTURED PATCHES WITH REGION ANNOTATION
Emit each actionable fix as a structured patch: target (CSS-ish selector hint, e.g. section#hero h1), bbox (normalized [x,y,w,h] in [0,1]; omit when no screenshot or page-wide), current_value (optional snippet), suggested_value, op (replace | add | remove | style | move; style for class/CSS-only changes), rationale (one sentence visual reason), severity (low | medium | high).

## USER MESSAGE (filled at runtime)
{verbatim_brief_block}
Review round: {round_idx}
Rendered screenshot path: {screenshot_path or "(missing)"}
Reference screenshot path: {reference_screenshot_path or "(missing)"}
{persistence_block}
Brand brief: {brand_brief_json}
Raw HTML: {html_doc}
[If screenshot present] Below: rendered screenshot of the page being reviewed. <image_part>
...(continue)...
\end{lstlisting}
\end{figure*}

\begin{figure*}[!p]
\begin{lstlisting}[
  basicstyle=\small\ttfamily,
  breaklines=true,
  frame=single,
  caption={Prompt for MessageCritic},
  columns=fullflexible,
  label={lst:message_critic_prompt},
  keepspaces=true,
  literate={-}{-}1
]
## ROLE
You are a message critic for conversion-focused landing pages.

## TASK
Review the HTML as text. Score tone_fidelity, value_prop_clarity, audience_signaling, must_have_sections, and claim_grounding from 1 to 5. claim_grounding measures how well every factual claim in the HTML (metrics, customers, integrations, awards, partnerships, pricing, certifications, capability claims) is supported by the brand brief -- 5 = every claim traceable to the brief; 1 = the page invents claims. List each ungrounded claim as a separate `issues` entry prefixed `ungrounded:`. approved=true only when the value proposition is unambiguously communicated above the fold AND the page satisfies the brief without inventing unsupported claims.

## INTENTIONAL PLACEHOLDER CONVENTION
Obvious-placeholder tokens are deliberately inserted by the polisher when the brief lacks a real number and are non-claims for claim_grounding -- do NOT flag them. ...(continue)... IS an ungrounded claim and must be flagged.

## must_have_sections SCORING -- ABSENT FACT-DATA SECTIONS ARE NOT A FAILURE
A fact-data section (customer-logo strip, testimonials/reviews, social-proof/stat band, press, awards) is correctly removed by the polisher when the brief supplies no real content. Do NOT lower must_have_sections for its absence and do NOT emit a patch to add it back. Only penalize when a NON-fact section the brief clearly warrants (hero, features, how-it-works, cta, footer) is missing.

## INPUT DATA
- Brand brief JSON.
- Review-round HTML.
- Auto-harvested candidate factual claims extracted from the HTML.
- Optional patch-persistence report from the previous round.

## OUTPUT  (strict JSON, no markdown)
{
  "scores": {
    "tone_fidelity": <1-5>, "value_prop_clarity": <1-5>, "audience_signaling": <1-5>,
    "must_have_sections": <1-5>, "claim_grounding": <1-5>
  },
  "issues":      [<short strings; ungrounded claims prefixed `ungrounded:`>],
  "suggestions": [<short strings>],
  "patches": [
    { "target": <CSS-ish selector hint>, "current_value": <<=80-char snippet or null>,
      "suggested_value": <replacement copy, or one-line description for remove>,
      "op": replace | add | remove,
      "rationale": <one-sentence messaging reason>,
      "severity": low | medium | high }
  ],
  "approved": <bool>
}

## USER MESSAGE (filled at runtime)
{verbatim_brief_block}
Review round: {round_idx}
{persistence_block}
## CANDIDATE FACTUAL CLAIMS (auto-harvested from HTML)
For each entry, check brief support; if unsupported, emit an `ungrounded:` issue AND a `remove` patch. Non-exhaustive hint; still scan full HTML for unsupported factual or capability claims.
{candidate_claims_json}
Brand brief: {brand_brief_json}
HTML: {polisher_html_rN}
\end{lstlisting}
\end{figure*}

\begin{figure*}[!p]
\begin{lstlisting}[
  basicstyle=\small\ttfamily,
  breaklines=true,
  frame=single,
  caption={Prompt for FunctionalCritic},
  columns=fullflexible,
  label={lst:functional_critic_prompt},
  keepspaces=true,
  literate={-}{-}1
]
## ROLE
You are a functional critic for one-page landing sites. You audit whether navigation avoids off-page destinations and whether in-page anchors are wired or need advisory fixes.

## TASK
Read the HTML and verify the one-page site's navigation. Score three dimensions from 1 to 5: anchor_integrity, nav_coverage, cta_targeting. Emit free-text issues/suggestions and structured patches the polisher can apply. Set approved=true iff every score >= 4.

## INPUT DATA
- Brand brief JSON (use must_have_sections to judge nav coverage).
- Review round HTML.
- Deterministic ANCHOR AUDIT block (counts of internal/empty/broken anchors, external URLs, action links, available ids).

## OUTPUT  (strict JSON, no markdown)
{
  "scores": {
    "anchor_integrity": <1-5>, "nav_coverage": <1-5>, "cta_targeting": <1-5>
  },
  "issues":      [<short strings>],
  "suggestions": [<short strings>],
  "patches": [
    { "target": <CSS-ish selector hint>, "current_value": <short snippet or null>,
      "suggested_value": <what to change it to>,
      "op": replace | add | remove | rename,
      "rationale": <one sentence>, "severity": low | medium | high }
  ],
  "approved": <bool -- true iff every score >= 4>
}

## SCORING DIMENSIONS
- anchor_integrity -- computed DETERMINISTICALLY from len(audit.external). EXTERNAL urls (http(s)://, protocol-relative //, or relative paths to other pages) DO lower it: 5 = 0, 4 = 1-2, 3 = 3-5, 2 = 6-10, 1 = 11+. Empty `#` or broken `#X` internal anchors are ADVISORY ONLY. mailto:/tel: neutral.
- nav_coverage -- nav/header links collectively cover the page's main sections. 5 = every meaningful section reachable from nav; 1 = nav skips the page's primary sections.
- cta_targeting -- primary and secondary CTAs point to an in-page anchor (`#section-id`) or legit contact action (mailto:/tel:). External destinations count AGAINST this score. A dead `#` is acceptable here (advisory). 5 = all CTAs stay in-page; 1 = primary CTA is external.

## STRUCTURED PATCHES GUIDANCE
Emit each actionable fix as a structured patch: target (CSS-ish selector hint, e.g. `nav a:nth-of-type(2)`), current_value (optional), suggested_value (e.g. `href="#features"`), op (replace | add | remove | rename), rationale (one sentence), severity (low | medium | high; high = external URL that should be in-page anchor; medium = missing/weak nav coverage; low = empty `#` or broken `#X` -- advisory, do not dock the score). Keep issues/suggestions short; specifics live in patches.

## USER MESSAGE (filled at runtime)
{verbatim_brief_block}
Review round: {round_idx}
{persistence_block}
## DETERMINISTIC ANCHOR AUDIT (ground truth, already computed)
- total hrefs: {audit.total}
- internal anchors -- empty: {audit.internal_empty}, broken: {len(audit.internal_broken)}, resolved: {len(audit.internal_resolved)}
- EXTERNAL urls: {len(audit.external)} -- examples: {audit.external[:8]}
- action links (mailto/tel): {audit.action_links}
- ids available: {audit.ids[:20]}

SCORING POLICY for anchor_integrity (computed from len(audit.external) per the table above). Your remaining job: judge nav_coverage and cta_targeting (semantic).

Brand brief: {brand_brief_json}
HTML: {polisher_html_r{round_idx}}
\end{lstlisting}
\end{figure*}

\renewcommand{\theequation}{C.\arabic{equation}}
\newpage
\section{Evaluation Protocol Details}
\label{app:evaluation_details}

\subsection{Metric Definitions}
\label{app:metric_definitions}

We evaluate generated landing pages using five metrics: Diversity, Faithfulness, Conciseness, Readability, and Aesthetics.
The definitions for each metric are compiled in \autoref{tab:metric_definitions}.

\begin{table*}[h]
\centering
\small
\begin{tabularx}{\linewidth}{lL}
\toprule
\textbf{Metric} & \textbf{Definition} \\
\midrule
Diversity
& Measures whether outputs generated for the same landing target differ in section composition, information flow, and layout pattern rather than repeating the same template. \\

\midrule

Faithfulness
& Measures whether the generated page accurately reflects the target specification, including product or service description, value proposition, intended audience, and conversion goal. \\

Conciseness
& Measures whether the page communicates essential information without unnecessary repetition, excessive explanation, or vague promotional copy. \\

Readability
& Measures whether the page has clear section flow, legible text, identifiable CTAs, and an interpretable layout. \\

Aesthetics
& Measures the overall visual completeness of the page, including color use, typography, spacing, alignment, visual balance, and perceived polish. \\
\bottomrule
\end{tabularx}
\caption{Evaluation metric definitions.}
\label{tab:metric_definitions}
\end{table*}


\subsection{Diversity Metric}
\label{app:diversity_metric}

Diversity is computed \emph{within} outputs generated from the same landing target, so that the score reflects variation induced by the generation method rather than differences among target briefs. For each landing target \(b \in \mathcal{B}\) and method \(m\), we generate \(K{=}3\) pages ($\{y_{b,m}^{(k)}\}_{k=1}^{K}$), render each to a full-page RGB screenshot, embed it with a frozen self-supervised vision encoder, and aggregate the pairwise cosine distances across the \(K\) outputs.

\paragraph{Preprocessing.}
Each generated page \(y\) is rendered to a full-page RGB screenshot \(S(y) \in [0,255]^{H \times W \times 3}\) at native width and full scroll height. Landing pages are vertically long (\(H \gg W\)), so a direct anisotropic resize to a square would compress vertical layout structure by an order of magnitude. We instead rescale \(S(y)\) by \(s = \tau / W\), preserving its aspect ratio, and crop the top-aligned \(\tau {\times} \tau\) square (\(\tau = 518\)):
\begin{equation}
c(y)=\mathrm{Crop}_\tau\!\bigl(\mathrm{Resize}_s(S(y))\bigr) \in  \mathbb{R}^{\tau \times \tau \times 3}.
\end{equation}

For landing pages the rescaled width equals \(\tau\), so the crop spans the full width and captures the rendered region from row \(0\) to row \(\tau/s\) in original pixels --- the above-the-fold and immediately scrollable hero region, which carries the dominant layout signal.

\paragraph{Visual encoder.}
We embed \(c(y)\) with the frozen DINOv3 ViT-L/16 backbone~\cite{eval2025dinov3} to obtain the page-level embedding:
\begin{equation}
\phi(y) \;=\; \mathrm{DINOv3}\bigl(c(y)\bigr).
\end{equation}
The encoder is not fine-tuned, and the same weights and preprocessing are applied to every method's outputs.

\paragraph{Pairwise distance.}
Pages are compared by cosine distance between L2-normalized embeddings:
\begin{equation}
D(y_i, y_j) \;=\; 1 \;-\; \frac{\langle \phi(y_i),\,\phi(y_j) \rangle}{\|\phi(y_i)\|_2 \cdot \|\phi(y_j)\|_2}.
\end{equation}
The distance is symmetric and equals zero only when the two embeddings point in the same direction.

\paragraph{Aggregation.}
The diversity score ($\mathcal{D}_s$) is the mean pairwise distance over all output pairs per target, averaged over targets:
\begin{equation}
\begin{split}
\mathrm{Div}(m) &= \frac{1}{|\mathcal{B}|} \sum_{b \in \mathcal{B}} \frac{1}{K(K-1)} \\
&\quad \times \sum_{1 \leq i < j \leq K} D\!\bigl(y_{b,m}^{(i)}, y_{b,m}^{(j)}\bigr).
\end{split}
\end{equation}

Higher values indicate more visually varied outputs for the same target.







\subsection{LLM-as-a-Judge Evaluation}
\label{app:laaj_evaluation}

LLM-as-a-Judge evaluation is used for Faithfulness, Conciseness, Readability, and Aesthetics. Each generated page is evaluated against the corresponding target specification. The judge receives the target specification and the rendered/generated result, and returns a score for each metric.

\begin{table}[h]
\centering
\small
\begin{tabularx}{\linewidth}{lL}
\toprule
\textbf{Item} & \textbf{Setting} \\
\midrule
Judge model & Claude Sonnet 4.6 \\
Evaluation format & Pointwise scoring \\
Score scale & 1--5 \\
Input to judge & Target specification and generated landing page. \\
Aggregation & Mean score per metric and method. \\
\bottomrule
\end{tabularx}
\caption{LLM-as-a-Judge evaluation settings.}
\label{tab:laaj_settings}
\end{table}

\subsection{LLM-as-a-Judge Rubric}
\label{app:laaj_rubric}

Table~\ref{tab:laaj_rubric} reproduces the 1--5 scoring rubric provided to the judge model for Faithfulness, Conciseness, Readability, and Aesthetics.

\begin{table*}[h]
\centering
\small
\begin{tabularx}{\textwidth}{llL}
\toprule
\textbf{Dimension} & \textbf{Score} & \textbf{Rubric} \\
\midrule
\multirow{5}{*}{Faithfulness}
  & 1 & Misrepresents the brand or omits the value proposition; fabricated claims. \\
  & 2 & Major gaps --- several sections the brief calls for missing, audience/tone clearly off, or only a generic brand mood with no concrete explanation of the offering. \\
  & 3 & Broadly on-brief but with notable omissions, vague substance (purpose/value not concretely conveyed), missing brief-supplied pricing/product detail, or one fabricated element. \\
  & 4 & Faithfully covers the brief; minor omissions only. \\
  & 5 & Fully faithful --- value prop, audience, tone, the concrete substance of the offering, brief-supplied pricing/product detail, and the sections the brief calls for all accurately represented. \\
\midrule
\multirow{5}{*}{Conciseness}
  & 1 & Severe walls of text, or near-empty of real content --- either way the message fails to land. \\
  & 2 & Consistently thin and generic, OR genuinely verbose with several aimless redundant blocks. \\
  & 3 & Adequate but with noticeable verbosity, an aimless redundant section, or copy thin enough to weaken the message. \\
  & 4 & Focused and substantive; minor tightening possible. \\
  & 5 & Every element earns its place --- substantive and focused, with repetition (if any) used purposefully. A rich, dense page belongs here when it stays focused. \\
\midrule
\multirow{5}{*}{Readability}
  & 1 & Illegible or chaotic --- broken hierarchy, unreadable text, overlap. \\
  & 2 & Frequent readability problems --- weak hierarchy, low contrast, disjointed section flow, or a hard-to-find CTA. \\
  & 3 & Readable but with weaknesses in hierarchy, narrative flow, section connection, or CTA prominence. \\
  & 4 & Clean and easy to scan; minor issues only. \\
  & 5 & Effortless to read and navigate; exemplary hierarchy, narrative flow, and a clear, prominent CTA. \\
\midrule
\multirow{5}{*}{Aesthetics}
  & 1 & Amateurish or broken styling; clashing colors; artifacts. \\
  & 2 & Weak, inconsistent design --- dated/default look, unstable alignment, a sparse/unfinished look, a flat hero, or monotonous throughout. \\
  & 3 & Acceptable but unremarkable --- inconsistencies in color/type/spacing, a thinly populated feel, a weak hero, or noticeable monotony. \\
  & 4 & Polished and professional, with a solid hero and some visual variety; minor refinements possible. \\
  & 5 & Exceptional, cohesive, top-tier visual design --- complete, varied, with a striking hero and publication-ready finish. \\
\bottomrule
\end{tabularx}
\caption{LLM-as-a-Judge scoring rubric.}
\label{tab:laaj_rubric}
\end{table*}

\subsection{LLM-as-a-Judge Prompt}
\label{app:laaj_prompt}

The condensed prompt template provided to the judge model is reproduced in Listing~\ref{lst:eval_prompt}. Each per-dimension \textsc{Definition} is preserved verbatim; the \textsc{What to check} bullets and the 1--5 rubric bands are abbreviated with ``\texttt{...(continue)...}'' since the full rubric bands are already reproduced in Table~\ref{tab:laaj_rubric}.

\begin{figure*}[!t]
\begin{lstlisting}[
  basicstyle=\small\ttfamily,
  breaklines=true,
  frame=single,
  caption={Prompt for LLM-as-a-Judge Evaluation},
  columns=fullflexible,
  label={lst:eval_prompt},
  keepspaces=true,
  literate={-}{-}1
]
## ROLE
You are an expert judge of landing-page quality. Score a SINGLE landing page on the {metrics_count} dimension(s) below: {metric_names}.
Judge each dimension INDEPENDENTLY against its own absolute rubric -- a weakness in one dimension must not pull down the score of another. Use the same page (HTML and/or screenshot) provided for every dimension.

## DIMENSIONS
The block below is composed at runtime from the per-dimension rubrics in DIMENSION RUBRIC LIBRARY. Only the dimensions enabled for this judge call appear, in canonical order (faithfulness -> conciseness -> readability -> aesthetics).
{criteria}

## DIMENSION RUBRIC LIBRARY (canonical text inserted into `{criteria}`)

### Faithfulness
**Definition** -- How accurately the page reflects the brand brief. The brief is the complete brand document the user supplied, NOT a page spec -- the page is not expected to surface every background/strategy detail it contains (market size, competitive moat, etc.). Judge two things: (a) essential, required brand information is NOT omitted from the page, and (b) NO false or contradictory information is added that the brief does not support. ...(continue)...

### Conciseness
**Definition** -- The signal-to-noise ratio of the page: every element conveys real information or moves the visitor forward. Conciseness is INFORMATION DENSITY and focus -- NOT word count, page length, or section count. ...(continue)...

### Readability
**Definition** -- How easily a visitor can scan, follow, and navigate the page: clear visual hierarchy, legible text, a logical reading flow, a coherent narrative across sections, usable navigation, adequate contrast. ...(continue)...

### Aesthetics
**Definition** -- Visual polish and design maturity: harmonious color, consistent typography, balanced spacing, modern professional styling that meets the standard of a well-funded company's site -- and a confident first impression. ...(continue)...

## OUTPUT FORMAT (Strict JSON)
Return ONLY a JSON object, no prose around it. It MUST contain an entry for every dimension listed in the ROLE header. The shape below is composed at runtime from the enabled dimensions in canonical order.
{ {json_fields} }

### `{json_fields}` shape per dimension (filled at runtime)
For every enabled metric `<m>` in canonical order, one entry of the form:
    "<m>": {"reasoning": "<2-4 sentences: what you observed, which rubric band it falls in, and the deciding factor>", "score": <integer 1-5>}
...(continue: worked-out example with all four dimensions enabled)...

## USER MESSAGE TEMPLATE (filled at runtime)
The user turn is assembled in `_build_combined_content` (see `utils/landing_evaluation.py`). The judge always receives a TARGET CONTEXT preface; the brand brief, page HTML / text, and rendered screenshot are appended only when the enabled metrics need them and the artifact is actually available.
{target_context_text}
...(continue: conditional artifact blocks for brand_brief_text / page_html / page_text / screenshot image_part, missing-artifact fallback notes, and `target_context_text` variants for target_kind in {"final","wireframe"})...
\end{lstlisting}
\end{figure*}

\subsection{Evaluation Target Coverage}
\label{app:evaluation_target_coverage}

The 30 target specifications span 12 industry categories, as summarized in Table~\ref{tab:evaluation-target-categories}.
They also vary in intended audience, CTA goal, and page style, covering consumer-facing pages, enterprise and product-led SaaS pages, community platforms, content-subscription pages, and infrastructure and product-catalog pages.
Each evaluation target has category-level candidate references in LandingBench, while the Retriever additionally matches references along structural and visual axes rather than industry category alone.

\begin{table}[h]
\centering
\small
\begin{tabular}{lrr}
\toprule
\textbf{Category} & \textbf{Count} & \textbf{\%} \\
\midrule
DevTools / Infra       & 3 & 10.0 \\
Media / Content        & 7 & 23.3 \\
Security               & 1 &  3.3 \\
B2B SaaS               & 4 & 13.3 \\
Marketing / Analytics  & 3 & 10.0 \\
Other                  & 1 &  3.3 \\
AdTech                 & 2 &  6.7 \\
Education              & 2 &  6.7 \\
E-commerce             & 1 &  3.3 \\
Social / Community     & 3 & 10.0 \\
Gaming                 & 1 &  3.3 \\
Health / Bio           & 2 &  6.7 \\
\midrule
\textbf{Total}         & \textbf{30} & \textbf{100.0} \\
\bottomrule
\end{tabular}
\caption{Industry-category distribution of the 30 evaluation target specifications. Percentages are rounded to one decimal place.}
\label{tab:evaluation-target-categories}
\end{table}

\subsection{Runtime and Reproducibility Settings}
\label{app:runtime_settings}

All generation experiments use a temperature of 0.7 where the model API supports temperature control. We do not specify a fixed generation seed. Instead, repeated generations are sampled stochastically under the same configuration. The prompt suite used in the reported experiments is designated version~1.0. Table~\ref{tab:runtime-settings} summarizes these settings.

\begin{table}[h]
\centering
\small
\begin{tabularx}{\linewidth}{lL}
\toprule
\textbf{Setting} & \textbf{Value} \\
\midrule
Temperature & 0.7 where supported by the model API \\
Generation seed & No fixed seed specified \\
Prompt-suite version & 1.0 \\
Maximum wireframe restarts & $R_{\max}=3$ \\
Maximum polishing rounds & $T_{\max}=3$ \\
\bottomrule
\end{tabularx}
\caption{Generation and reproducibility settings.}
\label{tab:runtime-settings}
\end{table}

Table~\ref{tab:iteration-distributions} reports the realized restart and retry counts across the 90 full-pipeline runs for the 30 evaluated targets. A count of zero indicates that the corresponding review loop did not require a repeated attempt. These counts describe control-loop invocation and are not standalone API or model failure rates.

\begin{table}[h]
\centering
\small
\begin{tabular}{lrrrr}
\toprule
\textbf{Count per run} & \textbf{0} & \textbf{1} & \textbf{2} & \textbf{3} \\
\midrule
Wireframe restarts & 71 & 9 & 4 & 6 \\
Polishing-review retries & 25 & 17 & 9 & 39 \\
\bottomrule
\end{tabular}
\caption{Observed restart and retry distributions across 90 full-pipeline runs.}
\label{tab:iteration-distributions}
\end{table}

Across these 90 full-pipeline runs, LandingAgent used approximately 428{,}000 tokens per generated page on average.

\subsection{Ablation Settings}
\label{app:ablation_settings}

The ablation study uses seven target specifications. For each condition, we generate three outputs per specification, resulting in 21 samples per condition and 126 samples across six conditions.

\begin{table}[h]
\centering
\small
\begin{tabularx}{\linewidth}{lL}
\toprule
\textbf{Variant} & \textbf{Description} \\
\midrule
Hard filter only
& Uses a retrieval pool constructed from pages that pass only hard filtering, without soft filtering or human verification. \\

w/o LandingBench
& Uses 438 samples randomly selected from Common Crawl as the RAG reference pool, preserving the retrieval mechanism and reference-pool size while replacing LandingBench. \\

w/o RAG
& Removes LandingBench-based retrieval from the generation process. \\

w/o Profiling
& Omits the Profiling phase from the generation process. \\

w/o Polishing
& Omits the final Polishing phase and uses the generated page draft as the final output. \\

Ours
& Uses the full LandingAgent pipeline with LandingBench-based retrieval and full filtering. \\
\bottomrule
\end{tabularx}
\caption{Ablation variants.}
\label{tab:ablation_variants}
\end{table}


\renewcommand{\theequation}{D.\arabic{equation}}
\section{Additional Qualitative Examples}
\label{app:more_qualitative}

All reported qualitative results were generated from specifications of real-world brands and subsequently de-identified for reporting. Brand-specific names and UI marks are replaced with fictional counterparts (e.g., Discord $\rightarrow$ Concord), and any remaining identifiable text is blurred.

\subsection{Comparison with Direct Prompting}

\begin{figure*}[h]
    \centering
    \includegraphics[width=\textwidth]{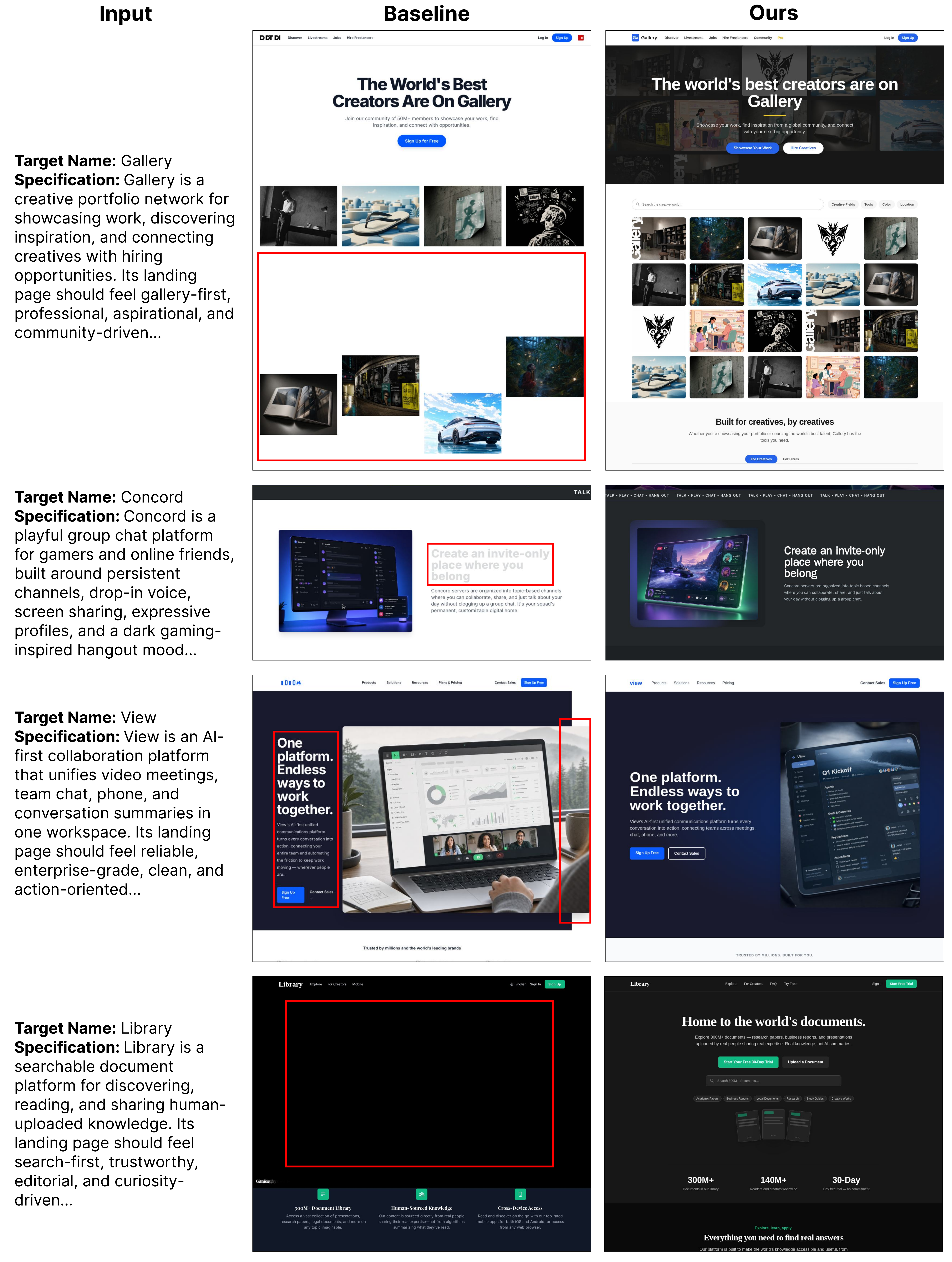}
    \caption{Additional qualitative comparisons between Direct Prompting and LandingAgent. From top to bottom: \textit{Gallery} --- the baseline omits required content; \textit{Concord} --- the baseline uses text whose color is too close to the background, hurting readability; \textit{View} --- the attached image overflows into the right margin and copy stretches vertically, reducing readability; \textit{Library} --- the baseline omits the CTA and renders an empty black region.}
    \label{fig:more-qualitative}
\end{figure*}

Figure~\ref{fig:more-qualitative} shows additional qualitative comparisons between Direct Prompting and LandingAgent across a broader set of target specifications.

\subsection{Brand De-identification}

\begin{figure*}[h]
    \centering
    \includegraphics[width=\textwidth]{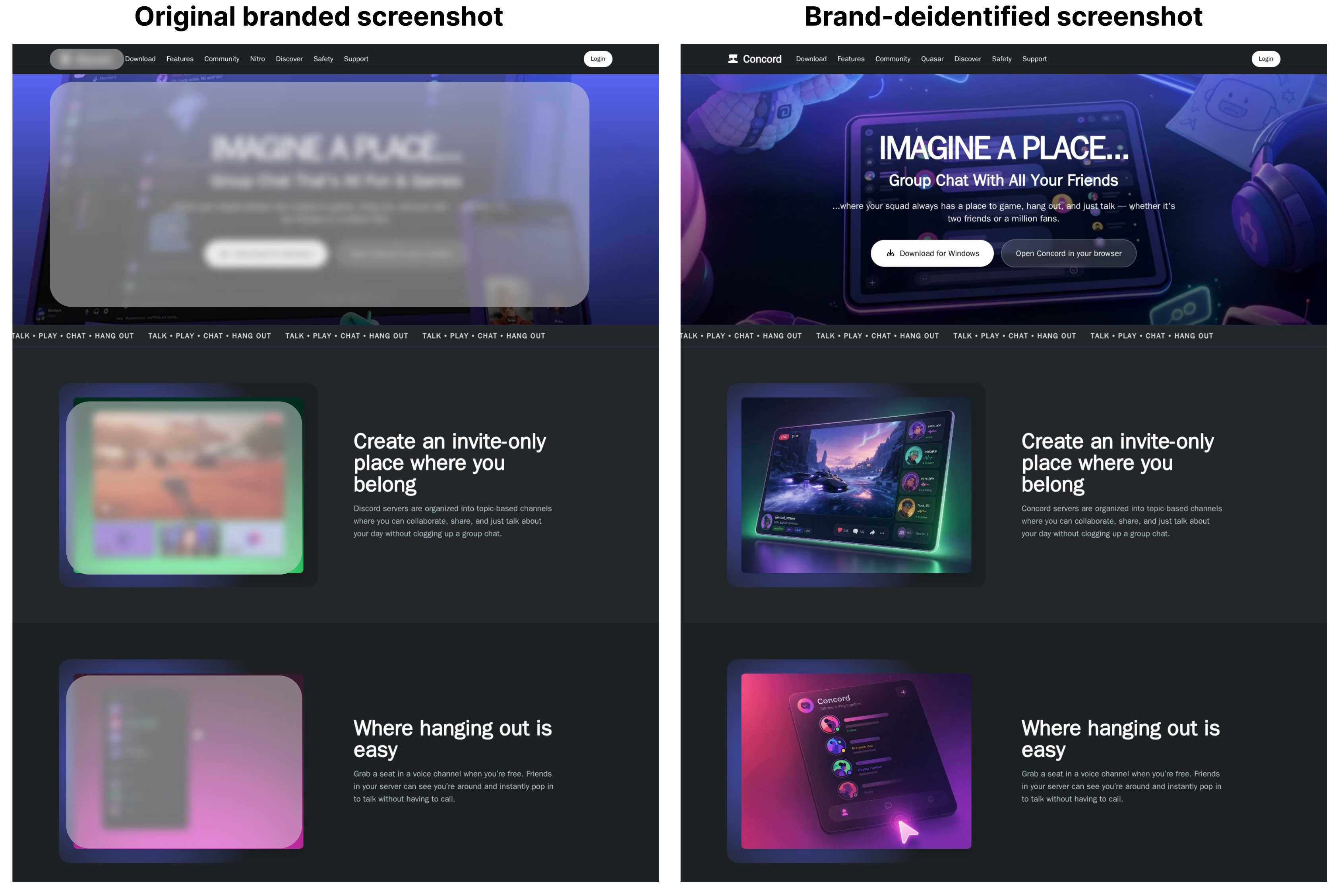}
    \caption{Example of brand de-identification for reporting. \textbf{Left:} Original screenshot, with identifiable text blurred to obscure recognizable commercial brand identifiers. \textbf{Right:} De-identified version in which all identifiable elements --- brand-specific names, UI marks, and other recognizable content --- are replaced with fictional counterparts (e.g., Discord $\rightarrow$ Concord).}
    \label{fig:deidentified}
\end{figure*}

Figure~\ref{fig:deidentified} illustrates the de-identification procedure used for all qualitative examples in this paper.

\end{document}